\documentclass[sigconf,nonacm]{acmart}
\usepackage{multirow}
\begin{document}

\title{CAAL: Confidence-Aware Active Learning for Heteroscedastic Atmospheric Regression}

\author{Fei Jiang}
\orcid{0009-0009-6697-9075}
\affiliation{%
  \department{Earth and Environmental Sciences}
  \institution{The University of Manchester}
  \city{Manchester}
  \state{Greater Manchester}
  \country{UK}
}
\email{fei.jiang@manchester.ac.uk}

\author{Jiyang Xia}
\orcid{0009-0003-4804-6382}
\affiliation{%
\department{Earth and Environmental Sciences}
  \institution{The University of Manchester}
  \city{Manchester}
  \state{Greater Manchester}
  \country{UK}
}
\email{jiyang.xia@manchester.ac.uk}

\author{Junjie Yu}
\orcid{0000-0003-4319-8988}
\affiliation{%
\department{Earth and Environmental Sciences}
  \institution{The University of Manchester}
  \city{Manchester}
  \state{Greater Manchester}
  \country{UK}
}
\email{junjie.yu@postgrad.manchester.ac.uk}

\author{Mingfei Sun}
\orcid{0000-0002-5925-5425}
\affiliation{%
\department{Department of Computer Science}
  \institution{The University of Manchester}
  \city{Manchester}
  \state{Greater Manchester}
  \country{UK}
}
\email{mingfei.sun@manchester.ac.uk}

\author{Hugh Coe}
\orcid{0000-0002-3264-1713}
\affiliation{%
\department{Earth and Environmental Sciences}
  \institution{The University of Manchester}
  \city{Manchester}
  \state{Greater Manchester}
  \country{UK}
}
\email{hugh.coe@manchester.ac.uk}

\author{David Topping}
\orcid{0000-0001-8247-9649}
\affiliation{%
\department{Earth and Environmental Sciences}
  \institution{The University of Manchester}
  \city{Manchester}
  \state{Greater Manchester}
  \country{UK}
}
\email{david.topping@manchester.ac.uk}

\author{Dantong Liu}
\orcid{0000-0003-3768-1770}
\affiliation{%
\department{Department of Atmospheric Sciences}
  \institution{Zhejiang University}
  \city{Hangzhou}
  \country{China}
}
\email{dantongliu@zju.edu.cn}

\author{Zhenhui Jessie Li}
\affiliation{%
  \institution{Yunqi Academy of Engineering}
  \city{Hangzhou}
  \country{China}
}
\email{jessielzh@gmail.com}

\author{Zhonghua Zheng}
\orcid{0000-0002-0642-650X}
\affiliation{%
\department{Earth and Environmental Sciences}
  \institution{The University of Manchester}
  \city{Manchester}
  \state{Greater Manchester}
  \country{UK}
}
\email{zhonghua.zheng@manchester.ac.uk}

\renewcommand{\shortauthors}{Jiang et al.}

\renewcommand\footnotetextcopyrightpermission[1]{}

\newcommand{\mingfei}[1]{\textcolor{red}{\{Mingfei: #1\}}}

\begin{abstract}
Quantifying the impacts of air pollution on health and climate relies on key atmospheric particle properties such as toxicity and hygroscopicity. However, these properties typically require complex observational techniques or expensive particle-resolved numerical simulations, limiting the availability of labeled data. We therefore estimate these hard-to-measure particle properties from routinely available observations (e.g., air pollutant concentrations and meteorological conditions). Because routine observations only indirectly reflect particle composition and structure, the mapping from routine observations to particle properties is noisy and input-dependent, yielding a heteroscedastic regression setting. With a limited and costly labeling budget, the central challenge is to select which samples to measure or simulate. While active learning (AL) is a natural approach, most acquisition strategies rely on predictive uncertainty. Under heteroscedastic noise, this signal conflates reducible epistemic uncertainty with irreducible aleatoric uncertainty, causing limited budgets to be wasted in noise-dominated regions. To address this challenge, we propose a confidence-aware active learning framework (CAAL) for efficient and robust sample selection in heteroscedastic settings. CAAL consists of two components: a decoupled uncertainty-aware training objective that separately optimises the predictive mean and noise level to stabilise uncertainty estimation, and a confidence-aware acquisition function that dynamically weights epistemic uncertainty using predicted aleatoric uncertainty as a reliability signal. Experiments on particle-resolved numerical simulations and real atmospheric observations show that CAAL consistently outperforms standard AL baselines. The proposed framework provides a practical and general solution for the efficient expansion of high-cost atmospheric particle property databases.

\end{abstract}

\keywords{Atmospheric science, Active learning, Heteroscedastic regression, Epistemic uncertainty, Aleatoric uncertainty}


\maketitle

\section{Introduction}
Air pollution significantly affects human health \cite{ZHANGETAL25Nature, LELIEVELDETAL15Nature} and climate change \cite{LIETAL16Nature, JACOBSON01Nature}, making it crucial to quantify its impacts comprehensively. However, properties of atmospheric particles, a key component of air pollution, vary dramatically in how easily they can be measured. Commonly observed properties, such as atmospheric particle concentrations and temperature, can be monitored through automated observation networks at relatively low cost. In contrast, some critical properties (e.g., toxicity \cite{HARKOV82ScienceofTheTotalEnvironment} and hygroscopicity \cite{WONETAL21SciRep}) require sophisticated observational techniques or intensive particle-resolved simulations and are therefore expensive and time-consuming to obtain \cite{POHLKERETAL23NatCommun, DAMIATIETAL25PublicHealthChallenges}. These hard-to-measure properties, however, more directly determine pollution impacts. For example, toxicity affects human health \cite{ZHENGETAL25Nature, DEDOUSSIETAL20Nature}, whereas hygroscopicity influences cloud formation and associated climate effects \cite{LIUETAL18NatCommun, ZHANGETAL23AtmosphericEnvironment, POHLKERETAL23NatCommun}. Developing methods to infer these hard-to-measure properties from routine observations would substantially advance our understanding of air pollution impacts.

A commonly used modelling framework for this inference task is heteroscedastic regression, which predicts both a mean and an input-dependent variance \cite{chaudhuri17, WANGJOSEPH25}, typically trained using a Gaussian negative log-likelihood (NLL) objective \cite{NIXWEIGEND94Proc.1994IEEEInt.Conf.NeuralNetw.ICNN94, KENDALLGAL17, LAKSHMINARAYANANETAL17, STIRNETAL22}. This matches the problem structure because the mapping from routinely observed inputs to hard-to-measure targets is inherently noisy and the noise level varies across conditions. The inputs include (1) easily measured pollution indicators (e.g., PM$_{2.5}$ concentrations) describing overall pollution levels and (2) contextual variables (meteorological conditions) describing environmental conditions. However, hard-to-measure properties are strongly influenced by particle-level chemical and physical heterogeneity \cite{HEALYETAL12AtmosphericChem.Phys., MARSHETAL19J.Phys.Chem.A, SCHWARZETAL06J.Geophys.Res., HUANGETAL24OneEarth}, which cannot be captured by routine observations. Consequently, similar inputs can correspond to very different particle-level properties and thus different target values.

While heteroscedastic regression is appropriate, the main bottleneck is data scarcity. Hard-to-measure labels are costly, requiring either labour-intensive particle-by-particle characterisation and aggregation over millions of particles, or regional particle-resolved simulations that demand petascale computing resources and tens of thousands of GPU cores \cite{HEALYETAL13Atmos.Chem.Phys., RIEMERETAL04Rev.Geophys., CURTISETAL24Geosci.ModelDev.}. Large-scale labelling is therefore impractical. Even multi-month or multi-year campaigns typically obtain only hundreds to thousands of regional samples \cite{HEALYETAL12AtmosphericChem.Phys., FADELETAL26AtmosphericPollutionResearch, DINGETAL25Environ.Sci.Technol.}. This is orders of magnitude smaller than routine monitoring archives such as OpenAQ, which has collected more than 25 billion air quality observations worldwide \cite{VIRAJSAWANTETAL}. This shifts the core problem to selecting the most informative samples to measure under a limited budget, motivating active learning (AL).

However, AL in heteroscedastic regression is challenging because predictive uncertainty has two distinct components \cite{HULLERMEIERWAEGEMAN21MachLearn, KENDALLGAL17}. Epistemic uncertainty (reducible uncertainty) reflects the model's incomplete knowledge due to limited training data \cite{KIUREGHIANDITLEVSEN09StructuralSafety}. Aleatoric uncertainty (irreducible uncertainty) reflects intrinsic variability in the input--output relationship \cite{KIUREGHIANDITLEVSEN09StructuralSafety}, which arises from unresolved particle-level heterogeneity in this application and cannot be reduced by collecting more data. Since both uncertainties can be high in the same region, selecting samples based on total uncertainty may waste limited resources on samples dominated by irreducible noise.

Previous approaches have attempted to address this issue either by discarding high-noise regions entirely \cite{ANTOSETAL10TheoreticalComputerScience} or by selecting samples solely based on epistemic uncertainty \cite{SHARMABILGIC17DataMinKnowlDisc, ABRAHAMDREYFUS-SCHMIDT22}. However, these strategies are overly extreme. Completely avoiding noisy regions may miss valuable learning opportunities, while ignoring noise levels can lead to queries dominated by aleatoric uncertainty with limited learning value. Furthermore, the NLL training objective couples mean and variance learning, allowing the model to reduce loss on noisy samples by simply predicting larger variances, thereby rescaling the residuals \cite{TAKAHASHIETAL18Proc.Twenty-SeventhInt.Jt.Conf.Artif.Intell., SEITZERETAL22, STIRNETAL22}. This can lead to unstable uncertainty estimates and unreliable epistemic uncertainty quantification, further undermining uncertainty-based AL.

In this work, we treat aleatoric uncertainty as an indicator of sample reliability for learning. High predicted noise suggests that epistemic uncertainty in that region is less reliable for guiding acquisition, whereas low predicted noise indicates that epistemic uncertainty more accurately reflects genuine learning opportunities. Based on this insight, we propose a Confidence-Aware Active Learning (CAAL) framework for heteroscedastic regression. Our framework consists of two key components. First, we employ a decoupled uncertainty-aware training objective to stabilise uncertainty estimation. The mean branch is trained with mean squared error (MSE), while the uncertainty branch independently models the residual scale using a Gaussian NLL. We introduce a weighting control mechanism to adjust the extent to which the uncertainty branch influences shared feature representations, preventing variance learning from destabilising mean learning. Second, we design a confidence-aware acquisition function that weights epistemic uncertainty by a confidence term derived from predicted aleatoric uncertainty, ensuring that selected samples offer both high learning potential and trustworthy uncertainty estimates.

Our main contributions can be summarised as follows:

\begin{itemize}
    \item A decoupled training objective for heteroscedastic regression that stabilises uncertainty estimation in AL settings.

    \item A confidence-aware acquisition function that balances epistemic and aleatoric uncertainty for robust sample selection.

    \item A practical framework for the cost-effective expansion of atmospheric particle property databases, with methodology applicable to other scientific domains where hard-to-measure properties must be inferred from low-cost observations under heteroscedastic conditions.
\end{itemize}

We evaluated CAAL on heteroscedastic atmospheric regression tasks using particle-resolved numerical simulations and real-world observations. On our primary particle-resolved simulation dataset, CAAL improve $R^2$ by 9.6$\%$ while using 45.6$\%$ fewer labels and it consistently outperforms standard AL baselines across additional datasets. The framework is particularly effective in highly heteroscedastic regions where standard methods struggle. This work takes a step towards building the data-rich foundations necessary for breakthrough predictive models of global air pollution, public health, and climate impact assessment.

\section{Related Work}
\subsection{Heteroscedastic Regression}
Heteroscedastic regression models input-dependent noise by predicting both the mean and variance, most commonly via a Gaussian NLL objective \cite{NIXWEIGEND94Proc.1994IEEEInt.Conf.NeuralNetw.ICNN94, KENDALLGAL17, LAKSHMINARAYANANETAL17, STIRNETAL22}. A known challenge is that, under heteroscedastic noise, the residual term in the NLL is scaled by the predicted variance, which can weaken optimisation pressure on the predictive mean in noise-dominated regions \cite{TAKAHASHIETAL18Proc.Twenty-SeventhInt.Jt.Conf.Artif.Intell., SEITZERETAL22, STIRNETAL22}.

Prior work has proposed several remedies that can be broadly grouped by how they handle this coupling. One line of research replaces the Gaussian likelihood with more robust heavy-tailed alternatives, such as the Student-$t$ distribution \cite{TAKAHASHIETAL18Proc.Twenty-SeventhInt.Jt.Conf.Artif.Intell.}, and further stabilises optimisation via additional parameterisations, marginalisation, or priors \cite{TAKAHASHIETAL18Proc.Twenty-SeventhInt.Jt.Conf.Artif.Intell., DETLEFSENETAL19, STIRNKNOWLES20}. A second line modifies the NLL through reweighting schemes, such as the $\beta$-NLL loss \cite{SEITZERETAL22}, to reduce the contribution of variance-related terms during optimisation. A third line explicitly alters gradient flow to decouple mean and variance learning by blocking variance-related gradients. Faithful \cite{STIRNETAL22} is a representative approach, where gradients from the variance (NLL) objective are prevented from updating the mean head and the shared feature extractor, ensuring faithful mean learning that is not degraded by variance optimisation. In addition, some works improve stability by reparameterising the heteroscedastic Gaussian likelihood using natural parameters, which changes how mean and variance interact in the objective and its gradients \cite{Alexander23}.

While these methods mainly target stable training and calibrated predictive uncertainty, heteroscedastic AL imposes an additional requirement on the uncertainty estimate because it is used to decide which samples to label. Approaches that retain strong coupling between the mean and variance may yield uncertainty estimates dominated by noise in difficult regions \cite{TAKAHASHIETAL18Proc.Twenty-SeventhInt.Jt.Conf.Artif.Intell., DETLEFSENETAL19, SEITZERETAL22, Alexander23}, whereas fully blocking variance-related gradients can stabilise mean learning but limit how noise structure influences representation learning \cite{STIRNETAL22}. This trade-off motivates controlled decoupling of the mean and variance, aiming to maintain stable mean learning while preserving uncertainty signals that remain useful for sample selection under heteroscedastic noise.

\subsection{Active Learning Acquisition}
AL aims to maximise model performance under a limited labelling budget by selecting the most informative samples for annotation \cite{KUMARGUPTA20J.Comput.Sci.Technol., RENETAL22ACMComput.Surv.}. In regression, a common baseline is uncertainty-based sampling, where samples with high predictive uncertainty are prioritised \cite{chaudhuri17, LANGE-HEGERMANNZIMMER25}. Entropy-based criteria maximise predictive entropy and are often referred to as Active Learning MacKay (ALM) \cite{MACKAY92NeuralComputation}. Disagreement-based methods measure predictive diversity across models or model instances, as in Query by Committee (QBC) \cite{SEUNGETAL92Proc.FifthAnnu.WorkshopComput.Learn.Theory, BURBIDGEETAL07IntelligentDataEngineeringandAutomatedLearning-IDEAL2007}, and are frequently used as benchmarks in regression AL \cite{CAIETAL132013IEEE13thInt.Conf.DataMin., WU18, WUETAL19InformationSciences, RIISETAL23}. Information-theoretic approaches select samples that maximise mutual information between predictions and model parameters, with BALD as a representative criterion targeting epistemic uncertainty \cite{ASHETAL20, RIISETAL23}. BatchBALD extends this idea to batch selection by accounting for redundancy \cite{KIRSCHETAL21}. BAIT incorporates representativeness constraints using information from the last-layer features \cite{Ashetal21}. 

In parallel, geometric and representativeness-based approaches select samples that cover the pool distribution in a learned feature space, as in coreset selection \cite{GEIFMANEL-YANIV17, SENERSAVARESE18}. BADGE constructs a gradient embedding for each unlabelled sample using the loss gradient with respect to the last-layer parameters, and then applies a k-means++-style batch selection procedure in this space, combining diversity with an uncertainty proxy induced by the current predictive distribution \cite{ASHETAL20}. ACS-FW connects batch acquisition to Bayesian coreset construction by using Frank-Wolfe optimisation to build a sparse subset that approximates the pool objective \cite{Pinslereyal2019}. LCMD is also clustering-based and shares a similar k-means++-style batch construction idea, aiming to form diverse and low-redundancy query sets \cite{HOLZMULLERETAL23}. 

For heteroscedastic regression, predictive uncertainty is commonly obtained using approximate Bayesian techniques such as Monte Carlo dropout \cite{KENDALLGAL17} or deep ensembles \cite{LAKSHMINARAYANANETAL17}, allowing it to be separated into epistemic and aleatoric components. Prior work suggests that identifying high-noise regions can reduce labelling costs \cite{chaudhuri17}. Existing acquisition strategies include avoiding high-noise regions \cite{ANTOSETAL10TheoreticalComputerScience}, adding confidence constraints \cite{chaudhuri17}, or focusing on reducible uncertainty \cite{SHARMABILGIC17DataMinKnowlDisc, ABRAHAMDREYFUS-SCHMIDT22}. However, these approaches typically treat these signals in isolation and do not explicitly balance reducible uncertainty against input-dependent noise in deep heteroscedastic regression.

\section{Preliminaries}
\subsection{Problem Setting}
We study pool-based AL for supervised regression.
Let $\mathbf{x} \in \mathbb{R}^d$ denote an input feature vector and $y \in \mathbb{R}$ the corresponding target.
We assume an unknown regression function $f:\mathbb{R}^d \rightarrow \mathbb{R}$ and a large unlabelled pool $\mathcal{U}$.
AL begins with an initial labelled set $\mathcal{D}^{(0)}_{L}=\{(\mathbf{x}_{i},y_{i})\}_{i=1}^{N_0}$, where $N_0$ is the initial labelling budget.
We denote the remaining unlabelled pool at round $t$ as $\mathcal{U}^{(t)}$.

AL proceeds for $T$ rounds with a fixed batch budget $B$ per round.
At round $t\in\{1,\dots,T\}$, a regression model is trained on the current labelled set $\mathcal{D}^{(t-1)}_{L}$.
An acquisition function $\alpha^{(t)}(\mathbf{x})$ is then evaluated for all candidates $\mathbf{x}\in\mathcal{U}^{(t-1)}$, and a query batch $\mathcal{Q}^{(t)}\subset\mathcal{U}^{(t-1)}$ of size $|\mathcal{Q}^{(t)}|=B$ is selected according to the acquisition scores.

After querying the oracle, the labelled and unlabelled sets are updated as
\begin{equation}
\mathcal{D}^{(t)}_{L}=\mathcal{D}^{(t-1)}_{L}\cup\{(\mathbf{x},y_{\mathrm{oracle}})\mid \mathbf{x}\in\mathcal{Q}^{(t)}\},
\qquad
\mathcal{U}^{(t)}=\mathcal{U}^{(t-1)}\setminus \mathcal{Q}^{(t)}.
\end{equation}
This procedure is repeated for $T$ rounds with query batch size $B$.

\subsection{Noisy Oracle and Heteroscedastic Data}
Unlike the noise-free AL setting, hard-to-measure labels are often noisy and input-dependent.
We adopt a Gaussian noise model as a standard, tractable approximation for continuous regression targets, which is widely used in heteroscedastic uncertainty modelling \cite{KENDALLGAL17}.
We model the oracle as follows:
\begin{equation}
y_{\mathrm{oracle}} = f(\mathbf{x})+\epsilon,
\qquad
\epsilon\sim \mathcal{N}\!\big(0,\sigma^2_{\mathrm{data}}(\mathbf{x})\big),
\end{equation}
where $\sigma^2_{\mathrm{data}}(\mathbf{x})$ denotes irreducible data uncertainty, e.g., intrinsic variability, measurement error, or modelling limitations.

\subsection{Heteroscedastic Regression Model}
Consistent with the oracle assumption, we adopt a Gaussian heteroscedastic regressor with parameters $\theta$.
\begin{equation}
p(y\mid \mathbf{x},\theta)=\mathcal{N}\!\big(y;\hat{\mu}_{\theta}(\mathbf{x}),\hat{\sigma}^2_{\theta}(\mathbf{x})\big).
\end{equation}
Both $\hat{\mu}_{\theta}(\mathbf{x})$ and $\hat{\sigma}^2_{\theta}(\mathbf{x})$ are parameterised using a shared feature extractor followed by two output heads (Section~\ref{sec:ensemble_backbone}).
Assuming conditional independence of labelled samples given the model parameters, we train the model by minimising the Gaussian NLL:
\begin{equation}
\mathcal{L}_{\mathrm{NLL}}(\theta)=
\frac{1}{N}\sum_{i=1}^{N}
\left(
\frac{1}{2}\log \hat{\sigma}^2_{\theta}(\mathbf{x}_i)
+
\frac{\big(y_i-\hat{\mu}_{\theta}(\mathbf{x}_i)\big)^2}{2\,\hat{\sigma}^2_{\theta}(\mathbf{x}_i)}
\right).
\end{equation}
In low-data regimes (typical in early AL rounds), joint NLL optimisation may reduce loss by inflating the predicted variance, which can weaken mean learning. This motivates our decoupled objective in Section~\ref{sec:mean_reg_loss}.

\section{Methodology}

\begin{figure*}
  \includegraphics[width=0.8\textwidth]{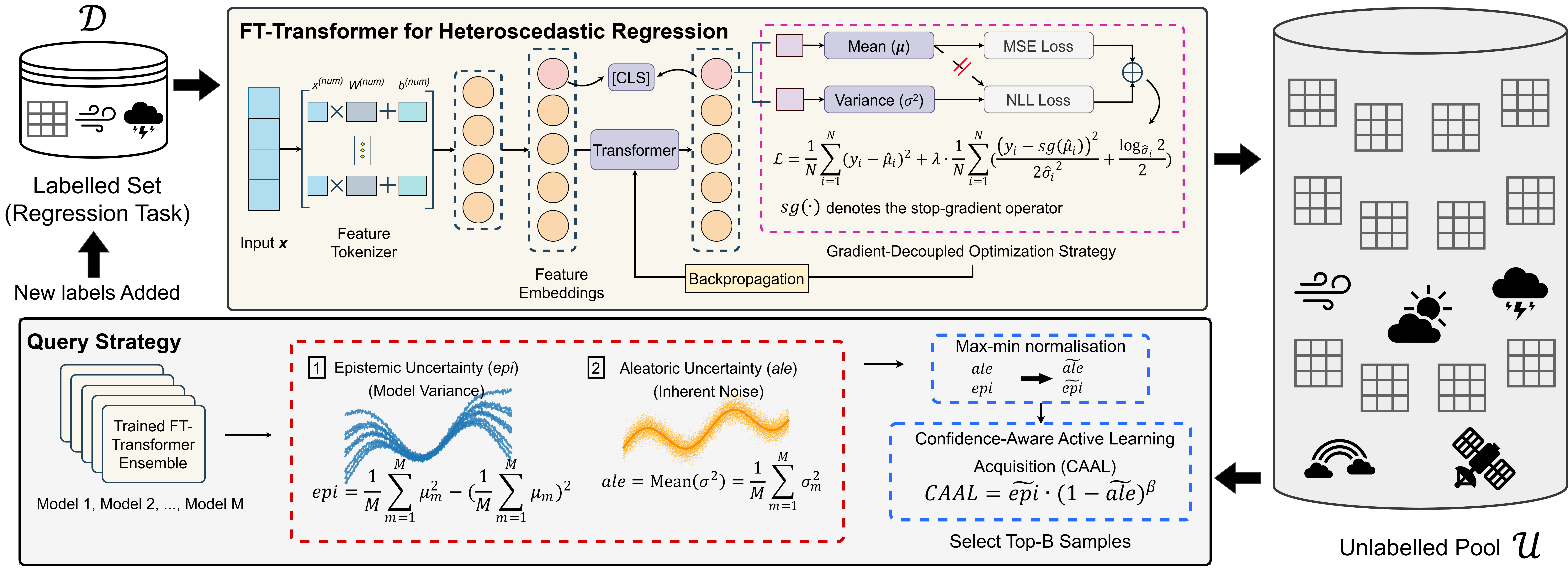}
  \caption{Overview of the Confidence-Aware Active Learning (CAAL) framework.}
  \label{fig:overview}
\end{figure*}

We propose the CAAL framework for heteroscedastic regression (Figure~\ref{fig:overview}).
CAAL has two explicitly defined components:
(1) a mean-variance decoupled training objective, and (2) a confidence-aware acquisition criterion.

\subsection{Uncertainty Estimation via Deep Ensembles}
\label{sec:ensemble_backbone}
We use deep ensembles of $M$ independently initialised heteroscedastic regressors, where $M$ is fixed and reported in the experimental setup.

For each ensemble member $m \in \{1,\dots,M\}$, the regressor consists of a shared feature extractor (trunk) followed by two task-specific heads. We denote its parameters by $\theta_m = (\theta^{\mathrm{trunk}}_m, \theta^{\mu}_m, \theta^{\sigma}_m)$ and use $\theta_m$ as a shorthand when there is no ambiguity. Given an input $\mathbf{x}$, the trunk produces a representation $\mathbf{h}_{m}(\mathbf{x})$, which is mapped to the predictive mean $\hat{\mu}_m(\mathbf{x})$ and an unconstrained variance logit $r_m(\mathbf{x})$ by the two heads:

\begin{equation}
\mathbf{h}_m(\mathbf{x}) = f_{\mathrm{trunk}}(\mathbf{x};\theta^{\mathrm{trunk}}_m),
\end{equation}

\begin{equation}
\hat{\mu}_m(\mathbf{x}) = g_{\mu}\!\left(\mathbf{h}_m(\mathbf{x});\theta^{\mu}_m\right),\qquad
r_m(\mathbf{x}) = g_{\sigma}\!\left(\mathbf{h}_m(\mathbf{x});\theta^{\sigma}_m\right),
\end{equation}

\begin{equation}
\hat{\sigma}^2_m(\mathbf{x}) 
= \log\!\left(1+e^{r_m(\mathbf{x})}\right) + \epsilon, \qquad
\hat{\sigma}^2_m(\mathbf{x}) > 0.
\end{equation}
Here, we set $\epsilon=10^{-6}$.

We decompose predictive uncertainty into the following two terms:
\textbf{Aleatoric uncertainty:}
\begin{equation}
\mathcal{U}_{\mathrm{ale}}(\mathbf{x})
=
\frac{1}{M}\sum_{m=1}^{M}\hat{\sigma}^2_{m}(\mathbf{x}),
\end{equation}

\textbf{Epistemic uncertainty:}
\begin{equation}
\mathcal{U}_{\mathrm{epi}}(\mathbf{x})
=
\frac{1}{M}\sum_{m=1}^{M}\hat{\mu}_{m}(\mathbf{x})^2
-
\left(
\frac{1}{M}\sum_{m=1}^{M}\hat{\mu}_{m}(\mathbf{x})
\right)^2.
\end{equation}

Under the standard mixture-of-Gaussians variance decomposition, the total predictive variance satisfies
\begin{equation}
\mathrm{Var}\!\left[y \mid \mathbf{x}, \mathcal{D}\right]
=
\mathcal{U}_{\mathrm{epi}}(\mathbf{x})
+
\mathcal{U}_{\mathrm{ale}}(\mathbf{x}).
\end{equation}

\subsection{Mean-Variance Decoupled Training}
\label{sec:mean_reg_loss}
Optimising a heteroscedastic likelihood can reduce loss by increasing predicted variance on difficult samples, which is especially problematic when labelled data are scarce in early AL rounds.
We therefore decouple mean learning and variance calibration at the gradient level.

For each ensemble member $m$, we optimise the following objective.

\textbf{Mean loss.}
We train the predictive mean using mean squared error (MSE):
\begin{equation}
\mathcal{L}_{\mu}(\theta_m)=
\frac{1}{N}\sum_{i=1}^{N}\big(y_i-\hat{\mu}_{m}(\mathbf{x}_i)\big)^2.
\end{equation}

\textbf{Variance calibration loss with stop-gradient residual.}
We calibrate predictive variance using an NLL-form term:
\begin{equation}
\mathcal{L}_{\sigma}(\theta_m)=
\frac{1}{N}\sum_{i=1}^{N}
\left(
\frac{1}{2}\frac{\big(y_i-\mathrm{sg}(\hat{\mu}_{m}(\mathbf{x}_i))\big)^2}{\hat{\sigma}^2_{m}(\mathbf{x}_i)}
+
\frac{1}{2}\log \hat{\sigma}^2_{m}(\mathbf{x}_i)
\right),
\end{equation}
where $\mathrm{sg}(\cdot)$ denotes the stop-gradient operator.

\textbf{Overall Objective.}
During backpropagation, $\mathcal{L}_{\mu}$ updates the $\theta^{\mathrm{trunk}}_m$ and $\theta^{\mu}_m$. For $\mathcal{L}_{\sigma}$, we use $\mathrm{sg}(\hat{\mu}_m)$ in the residual, so $\mathcal{L}_{\sigma}$ does not update the mean head parameters $\theta^{\mu}_m$. Gradients still update $\theta^{\mathrm{trunk}}_m$ and $\theta^{\sigma}_m$ via the variance branch. The strength of this interaction is controlled by $\lambda$.

We minimise the following objective for each ensemble member:
\begin{equation}
\mathcal{L}(\theta_m)
=
\mathcal{L}_{\mu}(\theta^{\mathrm{trunk}}_m,\theta^{\mu}_m)
+
\lambda\,\mathcal{L}_{\sigma}(\theta^{\mathrm{trunk}}_m,\theta^{\sigma}_m).
\end{equation}
Unless otherwise specified, we set $\lambda = 0.1$ in all experiments.
A detailed discussion of the role of $\lambda$ is provided in Section~\ref{sec:lambda_impact}.

\subsection{Baseline Training Objectives}
\label{sec:baseline_loss}

We compare the proposed training objective with three representative baselines for heteroscedastic regression.
All methods use the same model architecture, ensemble setting, and optimisation scheme.
Only the training objective or parameterisation is changed.

\textbf{$\beta$-NLL.}
The $\beta$-NLL baseline reweights the NLL using a variance-dependent weight with detached gradients \cite{SEITZERETAL22}:
\begin{equation}
\mathcal{L}_{\beta\text{-}\mathrm{NLL}}
=
\frac{1}{N}\sum_{i=1}^{N}
\big(\mathrm{sg}(\hat{\sigma}^2(\mathbf{x}_i))\big)^{\beta_{\mathrm{nll}}}
\left(
\frac{(y_i-\hat{\mu}(\mathbf{x}_i))^2}{2\,\hat{\sigma}^2(\mathbf{x}_i)}
+
\frac{1}{2}\log \hat{\sigma}^2(\mathbf{x}_i)
\right),
\end{equation}

where $\beta_{\mathrm{nll}}\in[0,1]$. We use $\beta_{\mathrm{nll}}=0.5$ and $1$ following \citeauthor{SEITZERETAL22}.

\textbf{Faithful.}
In the Faithful baseline, the NLL term is used solely for variance calibration and does not affect mean learning or representation learning \cite{STIRNETAL22}. The Faithful objective is

\begin{equation}
\mathcal{L}_{\mathrm{Faithful}}
=
\mathcal{L}_{\mu}(\theta^{\mathrm{trunk}}_m,\theta^{\mu}_m)
+
\mathcal{L}_{\sigma}(\theta^{\sigma}_m),
\qquad
\frac{\partial \mathcal{L}_{\sigma}}
{\partial \mathbf{h}_{m}(\mathbf{x})}
=0.
\end{equation}

\textbf{Natural Laplace.}
The Natural Laplace baseline predicts the natural parameters of a Gaussian distribution \cite{Alexander23} $(\eta_1(\mathbf{x}), \eta_2(\mathbf{x}))$, which are mapped to mean and variance by
\begin{equation}
\hat{\mu}(\mathbf{x}) = -\frac{\eta_1(\mathbf{x})}{2\eta_2(\mathbf{x})},
\qquad
\hat{\sigma}^2(\mathbf{x}) = -\frac{1}{2\eta_2(\mathbf{x})}, \qquad \eta_2(\mathbf{x})<0,
\end{equation}

\subsection{CAAL Acquisition}
\label{sec:acq}
To make $\mathcal{U}_{\mathrm{epi}}$ and $\mathcal{U}_{\mathrm{ale}}$ comparable, we apply min-max normalisation within the current unlabelled pool $\mathcal{U}^{(t-1)}$:
\begin{equation}
\tilde{\mathcal{U}}_{\star}(\mathbf{x})
=
\frac{
\mathcal{U}_{\star}(\mathbf{x})
-
\min_{\mathbf{x}'\in\mathcal{U}^{(t-1)}}\mathcal{U}_{\star}(\mathbf{x}')
}{
\max_{\mathbf{x}'\in\mathcal{U}^{(t-1)}}\mathcal{U}_{\star}(\mathbf{x}')
-
\min_{\mathbf{x}'\in\mathcal{U}^{(t-1)}}\mathcal{U}_{\star}(\mathbf{x}')
+\varepsilon
},
\end{equation}
where $\varepsilon=10^{-6}$ is a small constant for numerical stability.

At AL round $t$, CAAL scores each $\mathbf{x}\in\mathcal{U}^{(t-1)}$ by combining (i) its information potential ($\tilde{\mathcal{U}}_{\mathrm{epi}}(\mathbf{x})$) and (ii) its data confidence ($1-\tilde{\mathcal{U}}_{\mathrm{ale}}(\mathbf{x})$). Larger values indicate samples with lower aleatoric noise and therefore more reliable data for training.
The queried batch $\mathcal{Q}^{(t)}$ is selected as the top-$B$ points ranked by the CAAL acquisition score ($\alpha^{(t)}_{\mathrm{CAAL}}(\mathbf{x})$):
\begin{equation}
\alpha^{(t)}_{\mathrm{CAAL}}(\mathbf{x})=
\tilde{\mathcal{U}}_{\mathrm{epi}}(\mathbf{x})
\left(1-\tilde{\mathcal{U}}_{\mathrm{ale}}(\mathbf{x})\right)^{\beta},
\qquad
\beta\ge 0,
\end{equation}
Here, the exponent $\beta$ controls the strength of penalisation on candidates with high aleatoric uncertainty, balancing information seeking against data reliability. 
Unless otherwise specified, we set $\beta = 1$ in all experiments.
A detailed discussion of the role of $\beta$ is provided in Section~\ref{sec:beta_impact}.

\subsection{Baseline AL Acquisition Strategies}
\label{sec:baseline_acq}
All baselines use the same ensemble backbone (Section~\ref{sec:ensemble_backbone}), the same training objective (Section~\ref{sec:mean_reg_loss}), and the same uncertainty decomposition.
For geometry and clustering baselines, we operate on an embedding $\mathbf{z}(\mathbf{x})$ given by the layer-normalised CLS token output of the shared trunk. $\mathbf{z}(\mathbf{x})$ is obtained by averaging the CLS embeddings across ensemble members.

\textbf{Random Sampling.} Uniform sampling from $\mathcal{U}^{(t-1)}$.

\textbf{Confidence score.} 
We rank by
\begin{equation}
\alpha^{(t)}_{\mathrm{Confidence}}(\mathbf{x})=
1-\tilde{\mathcal{U}}_{\mathrm{ale}}(\mathbf{x}).
\end{equation}

\textbf{Ale.} 
We rank by
\begin{equation}
\alpha^{(t)}_{\mathrm{Ale}}(\mathbf{x})=
{\mathcal{U}}_{\mathrm{ale}}(\mathbf{x}).
\end{equation}

\textbf{ALM\cite{MACKAY92NeuralComputation}.}
Under a Gaussian approximation, predictive entropy is monotonic in total predictive variance.
We rank by
\begin{equation}
\alpha^{(t)}_{\mathrm{ALM}}(\mathbf{x})=
\mathcal{U}_{\mathrm{epi}}(\mathbf{x})+\mathcal{U}_{\mathrm{ale}}(\mathbf{x}).
\end{equation}

\textbf{QBC\cite{SEUNGETAL92Proc.FifthAnnu.WorkshopComput.Learn.Theory, BURBIDGEETAL07IntelligentDataEngineeringandAutomatedLearning-IDEAL2007}.} 
For regression ensembles, committee disagreement equals epistemic uncertainty.
\begin{equation}
\alpha^{(t)}_{\mathrm{QBC}}(\mathbf{x})=\mathcal{U}_{\mathrm{epi}}(\mathbf{x}).
\end{equation}

\textbf{BALD\cite{ASHETAL20, RIISETAL23}.} BALD selects samples by maximising the mutual information between the predictive output and the model parameters, conditioned on the input and the current labelled set.

\begin{equation}
\alpha^{(t)}_{\mathrm{BALD}}(\mathbf{x})
=
I\!\left(y, \boldsymbol{\theta} \mid \mathbf{x}, \mathcal{D}\right)
=
H\!\left[p\!\left(y \mid \mathbf{x}, \mathcal{D}\right)\right]
-
\mathbb{E}_{p(\boldsymbol{\theta}\mid\mathcal{D})}
\Big[
H\!\left[p\!\left(y \mid \mathbf{x}, \boldsymbol{\theta}\right)\right]
\Big],
\end{equation}

Under the Gaussian assumption, $H$ is monotonic in the predictive variance.
Dropping constants and positive scaling factors that do not affect ranking yields the ensemble-based approximation:

\begin{equation}
\alpha^{(t)}_{\mathrm{BALD}}(\mathbf{x})=
\frac{1}{2}\log\!\Big(\mathcal{U}_{\mathrm{epi}}(\mathbf{x})+\mathcal{U}_{\mathrm{ale}}(\mathbf{x})\Big)
-\frac{1}{2M}\sum_{m=1}^{M}\log\!\big(\hat{\sigma}^2_{m}(\mathbf{x})\big).
\end{equation}

\textbf{CoreSet\cite{GEIFMANEL-YANIV17, SENERSAVARESE18}.} 
Let $\mathbf{z}_i$ denote the embedding of sample $\mathbf{x}_i$. We construct a batch of size $B$ by greedy farthest-first traversal in $\mathbf{z}(\mathbf{x})$ space.
\begin{equation}
\mathcal{S}
=
\arg\min_{|\mathcal{S}|=B}
\max_{\mathbf{x}_i \in \mathcal{U}^{(t-1)}}
\min_{\mathbf{x}_j \in \mathcal{S}}
\lVert \mathbf{z}_i - \mathbf{z}_j \rVert_2.
\end{equation}

\textbf{LCMD\cite{HOLZMULLERETAL23}.} 
LCMD selects representative samples by clustering the unlabelled data in the embedding space.
We run $k$-means on $\{\mathbf{z}_i\}_{\mathbf{x}_i \in \mathcal{U}^{(t-1)}}$ with $k=B$, and select the sample closest to each cluster centroid.

\begin{equation}
\mathbf{x}_c
=
\arg\min_{\mathbf{x}_i \in \mathcal{C}_c}
\lVert \mathbf{z}_i - \boldsymbol{\mu}_c \rVert_2,
\end{equation}
where $\mathcal{C}_c$ and $\boldsymbol{\mu}_c$ denote the $c$-th cluster and its centroid.

\textbf{BADGE\cite{ASHETAL20}.}
BADGE combines uncertainty and diversity by constructing gradient-based embeddings and performing clustering in the resulting space.
In our regression setting, we use a proxy embedding based on the feature representation weighted by epistemic uncertainty.

\begin{equation}
\mathbf{g}_i
=
\mathbf{z}_i \cdot
\sqrt{\mathcal{U}_{\mathrm{epi}}(\mathbf{x}_i)}.
\end{equation}
We then run $k$-means with $k=B$ and $k$-means++ initialisation in the $\{\mathbf{g}_i\}$ space, and select the sample closest to each cluster centroid.

\section{Experimental Setup}

We compare CAAL with representative baselines along two axes, the training objective and the AL acquisition strategy.



\subsection{Dataset}
\textbf{Simulation data.} The PartMC dataset is generated using the particle-resolved aerosol model PartMC-MOSAIC \cite{RIEMERETAL09J.Geophys.Res.}, which serves as a benchmark for aerosol mixing state ($\chi$) estimation \cite{JIANGETAL25ACSESTAir}. Here, $\chi$ quantifies how uniformly chemical species are distributed across particles \cite{RIEMERWEST13Atmos.Chem.Phys., RIEMERETAL19ReviewsofGeophysics}. Simulation scenarios are constructed using Latin Hypercube Sampling (LHS) to ensure broad coverage of the input space, including environmental conditions, gas-phase emissions, and aerosol emission properties. Each scenario corresponds to a 24-hour simulation with hourly outputs, including the initial state at $t=0$ and the final state at $t=24$, yielding 25 samples per scenario. We consider two prediction targets derived from the same samples and input features: the chemical mixing state index $\chi_a$ and the optical mixing state index $\chi_o$ \cite{ZHENGETAL21EarthSpaceSci.}.

In total, 1,000 independent scenarios are generated (25,000 samples). To avoid temporal leakage, data are split at the scenario level, such that all hourly samples from the same scenario are assigned to the same split. The dataset is randomly divided into training (90$\%$), validation (5$\%$), and test (5$\%$) sets. Within the training set, 100 scenarios are randomly selected as the initial labelled set, while the remaining training scenarios are treated as the unlabelled pool for AL. 
AL is conducted at the scenario level for 20 rounds, querying 30 scenarios per round, which yields a fixed labelling budget of 600 scenarios in the particle-resolved simulation experiments. As an additional robustness check, we also run AL until the unlabelled pool is exhausted, with results reported in Appendix~\ref{append:alldata}. Further details on data generation and preprocessing are provided in Appendix~\ref{append:partmc}.

\textbf{Observational data.} We use hourly observational data of black carbon (BC) coating properties (VR) collected in the Beijing region from 2013 to 2019, measured using a Single Particle Soot Photometer (SP2) \cite{SCHWARZETAL06J.Geophys.Res.} in both urban and suburban environments \cite{HUETAL24Environ.Res.Lett.}. Here, VR describes how much coating material surrounds a BC core, which affects BC optical properties and its climate impact \cite{LIUETAL17NatureGeosci}.
The dataset is split chronologically. Data from 2013 are used as the initial labelled set (2,903 samples). Observations from 2014--2017 form the unlabelled pool (5,363 samples). Data from 2018 are used for validation (1,728 samples), and data from 2019 are reserved as the test set (1,208 samples). AL is conducted for 20 rounds, querying 250 samples per round from the unlabelled pool. Additional details on data preprocessing are provided in Appendix~\ref{append:VR}.

We conduct the main quantitative evaluation on the $\chi_a$ prediction task using particle-resolved simulation dataset. To assess stability and robustness, we further evaluate CAAL on additional targets $\chi_o$ from the same simulation dataset, and on VR prediction using the Beijing observational dataset.

\subsection{Baselines}

\textbf{Model training baselines.}
We consider standard heteroscedastic regression trained with Gaussian NLL, $\beta$-NLL, Faithful, and Natural Laplace.
Detailed formulations are provided in Section~\ref{sec:baseline_loss}.

\textbf{Active learning baselines.}
We compare CAAL with random sampling, uncertainty-based methods (ALM, Ale, Confidence, QBC, and BALD) and geometry- or representation-based methods (Coreset, BADGE, and LCMD).
Detailed formulations are provided in Section~\ref{sec:baseline_acq}.

\subsection{Implementation Details}
Models are implemented using PyTorch and trained on an NVIDIA RTX 5090 GPU. We adopt the FT-Transformer architecture \cite{GORISHNIYETAL23} as the regression backbone, setting the model dimension to $D_{\text{model}}=128$ with 8 attention heads, 4 transformer layers, and a feed-forward dimension of 512. The Gaussian Error Linear Unit (GELU) activation is applied throughout the network.

Models are trained using the Adam optimiser with a batch size of 128 for up to 100 epochs. Early stopping is applied when the validation loss does not improve for 20 consecutive epochs. The initial learning rate is set to 1e-4 and is reduced by a factor of 0.5 if the validation loss plateaus for 10 epochs, with a minimum learning rate of 1e-7. Hyperparameters are tuned with Optuna \cite{AKIBAETAL19} and then fixed across all active learning strategies and loss variants for fair comparison. A detailed configuration of the model architecture and training setup is provided in Appendix~\ref{sec:config}.

\section{Results}

\subsection{Overall Active Learning Performance}

\begin{figure}
  \centering
  \includegraphics[width=\linewidth]{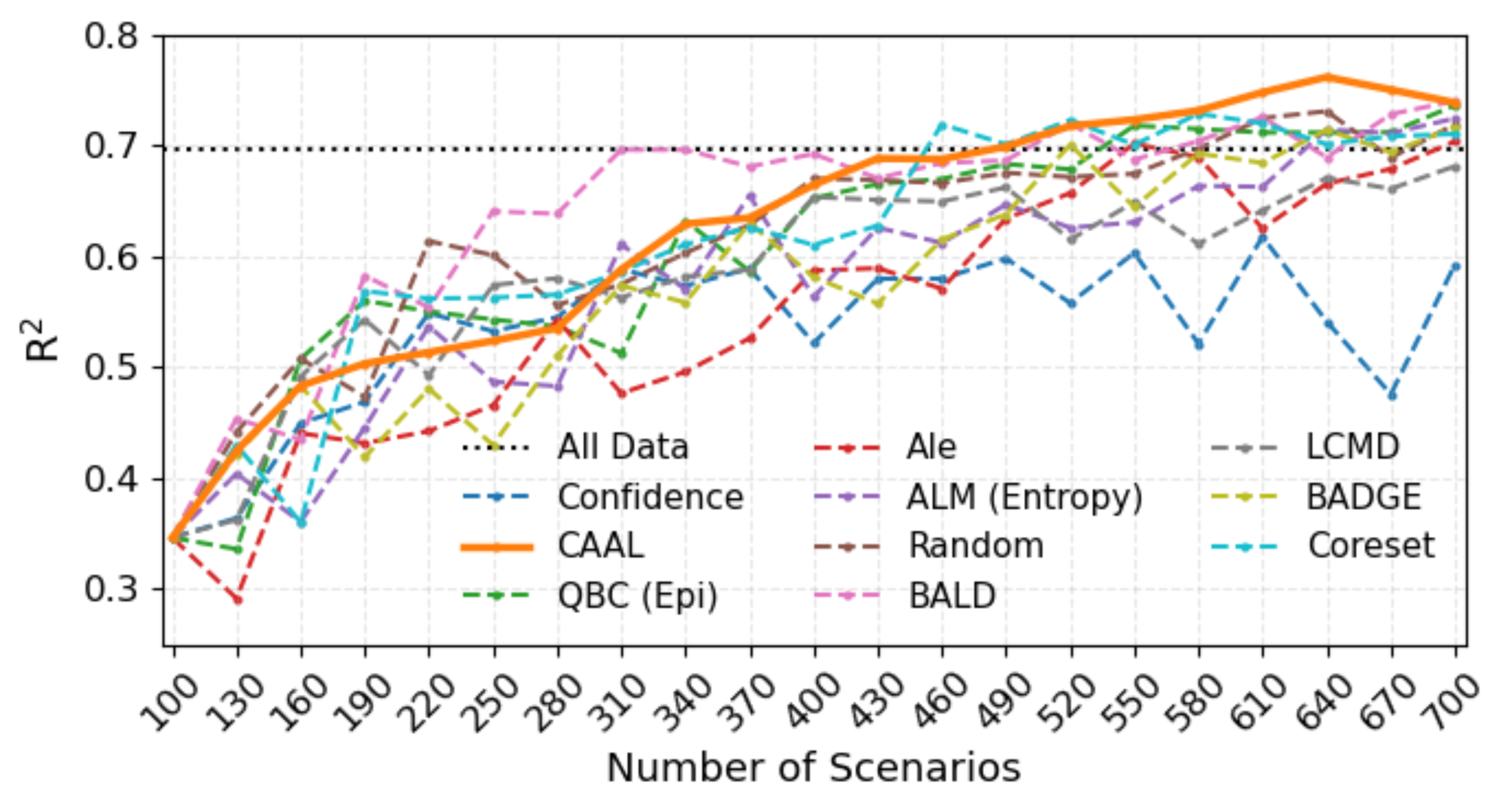}
  \caption{$R^2$ for $\chi_a$ on the PartMC dataset versus the AL query budget, evaluated at the scenario level.}
  \label{fig:r2_baseline}
\end{figure}

\begin{table}
\centering
\caption{Performance of AL strategies for $\chi_a$ on the PartMC dataset. Best $R^2$ and RMSE are shown with differences from full-data training. ``Labelling Saved'' denotes annotation cost reduced to match the performance of full-data training.}
\label{tab:main_results}
\resizebox{\linewidth}{!}{
\begin{tabular}{lcccccc}
\toprule
\multirow{2}{*}{\textbf{Method}} & \multicolumn{2}{c}{\textbf{Effectiveness ($R^2$)}} & \multicolumn{2}{c}{\textbf{Effectiveness (RMSE)}} & \multicolumn{2}{c}{\textbf{Efficiency}} \\
\cmidrule(lr){2-3} \cmidrule(lr){4-5} \cmidrule(lr){6-7}
 & Best $R^2$ ($\uparrow$) & vs. Full Data & Best RMSE ($\downarrow$) & vs. Full Data & Data to Match & \textbf{Labeling Saved} \\
\midrule
\textit{Full Data (Ref.)} & \textit{-} & \textit{0.6966} & \textit{-} & \textit{0.0728} & \textit{100\%} & \textit{0\%} \\
\midrule
Confidence & 0.6180 & -11.28\% & 0.0813 & +11.73\% & N/A & N/A \\
Ale & 0.7041 & +1.08\% & 0.0716 & -1.68\% & 61.1\% & 38.9\% \\
ALM (Entropy) & 0.7248 & +4.05\% & 0.0690 & -5.18\% & 71.1\% & 28.9\% \\
QBC (Epi) & 0.7365 & +5.73\% & 0.0676 & -7.22\% & 61.1\% & 38.9\% \\
BALD & 0.7406 & +6.31\% & 0.0670 & -7.93\% & 57.8\% & 42.2\% \\
LCMD & 0.6817 & -2.14\% & 0.0743 & +1.98\% &  N/A & N/A \\
BADGE & 0.7168 & +2.90\% & 0.0700 & -3.81\% & 57.8\% & 42.2\% \\
Coreset & 0.7108 & +4.67\% & 0.0685 & -5.93\% & \textbf{51.1\%} & \textbf{48.9\%} \\
Random & 0.7313 & +4.98\% & 0.0682 & -6.30\% & 67.8\% & 32.2\% \\
\textbf{CAAL (Ours)} & \textbf{0.7621} & \textbf{+9.40\%} & \textbf{0.0642} & \textbf{-11.83\%} & 54.4\% & 45.6\% \\
\bottomrule
\end{tabular}}
\label{tab:eff}
\end{table}

Figure~\ref{fig:r2_baseline} and Table~\ref{tab:eff} compare the performance of different AL strategies for $\chi_a$ on the PartMC dataset. Overall, CAAL exhibits a consistently stable improvement throughout the acquisition process and achieves the best overall performance among all methods, with a $R^2$ of 0.7621 and an RMSE of 0.0642. Compared with the full-data baseline, this corresponds to a 9.4$\%$ increase in $R^2$ and an 11.83$\%$ reduction in RMSE.

Among uncertainty-driven approaches, BALD performs competitively under the heteroscedastic regression. This acquisition uses the predictive output distribution, which contains both model uncertainty and data noise \cite{ASHETAL20, RIISETAL23}, but it does not explicitly separate epistemic and aleatoric uncertainty. BALD and QBC achieve strong gains in early acquisition rounds, yet exhibit reduced stability and inferior final performance compared with CAAL in later stages. In contrast, Total predictive uncertainty (ALM) and pure aleatoric-driven strategies (Ale and Confidence) show pronounced instability and limited performance gains. Geometry-based sampling (Coreset, LCMD, and BADGE) yields substantially lower final predictive accuracy than uncertainty-aware methods.

In terms of labelling efficiency, CAAL matches the full-data baseline using only 54.4$\%$ of the samples, corresponding to a 45.6$\%$ reduction in labelling cost, while achieving a relative $R^2$ improvement of 7.22$\%$ over Coreset, which yields the highest labelling savings.

\subsection{Why CAAL Works: A Virtuous Cycle}
\label{sec:mechanism}

\begin{figure}
  \centering\includegraphics[width=\linewidth]{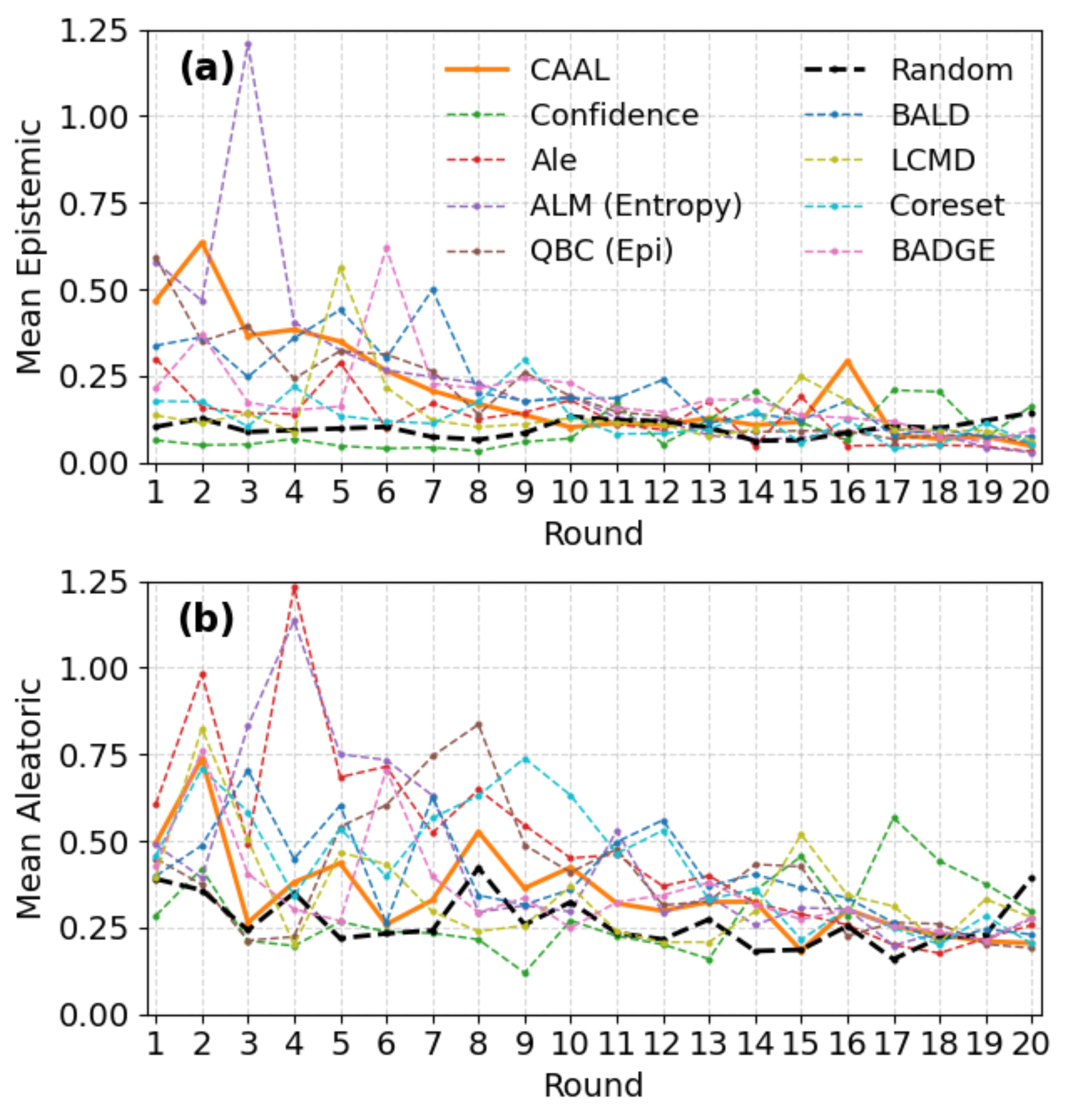}
  \caption{Mean epistemic uncertainty (a) and aleatoric uncertainty (b), averaged over the queried samples at each AL round on the PartMC dataset.}
  \label{fig:evolution}
\end{figure}

CAAL’s performance advantage arises from its decoupled treatment of epistemic and aleatoric uncertainty throughout the AL process. By guiding exploration with epistemic uncertainty in the early stage and maintaining effective noise control in the later stage, CAAL achieves consistent and robust sample selection across different learning phases.

Rather than aggressively maximising epistemic uncertainty, CAAL prioritises samples that exhibit high epistemic uncertainty while remaining well controlled for aleatoric noise, thereby balancing information gain against noise risk. This allows CAAL to acquire samples with epistemic uncertainty comparable to those selected by the Epi strategy (Figure~\ref{fig:evolution}a), while simultaneously maintaining substantially lower aleatoric uncertainty (Figure~\ref{fig:evolution}b), avoiding the premature inclusion of high-noise samples in the early AL stage.

As AL progresses, the overall epistemic uncertainty in the unlabelled pool naturally decreases, and sample selection becomes increasingly susceptible to irreducible noise. In the later AL stage, CAAL consistently maintains a controlled aleatoric uncertainty profile: the selected samples remain close to the random baseline in terms of noise level, while still retaining relatively high epistemic uncertainty (Figure~\ref{fig:evolution}). This behaviour effectively suppresses noise-dominated uncertainty and prevents the accumulation of high-noise samples that would otherwise degrade generalisation. Overall, CAAL forms a stable virtuous cycle: high-quality sample selection leads to more reliable uncertainty estimation, which in turn supports increasingly robust and effective sample selection in subsequent rounds.

\subsection{How \texorpdfstring{$\beta$}{beta} Controls Noise Tolerance in CAAL}
\label{sec:beta_impact}
\begin{figure}
  \centering
  \includegraphics[width=\linewidth]{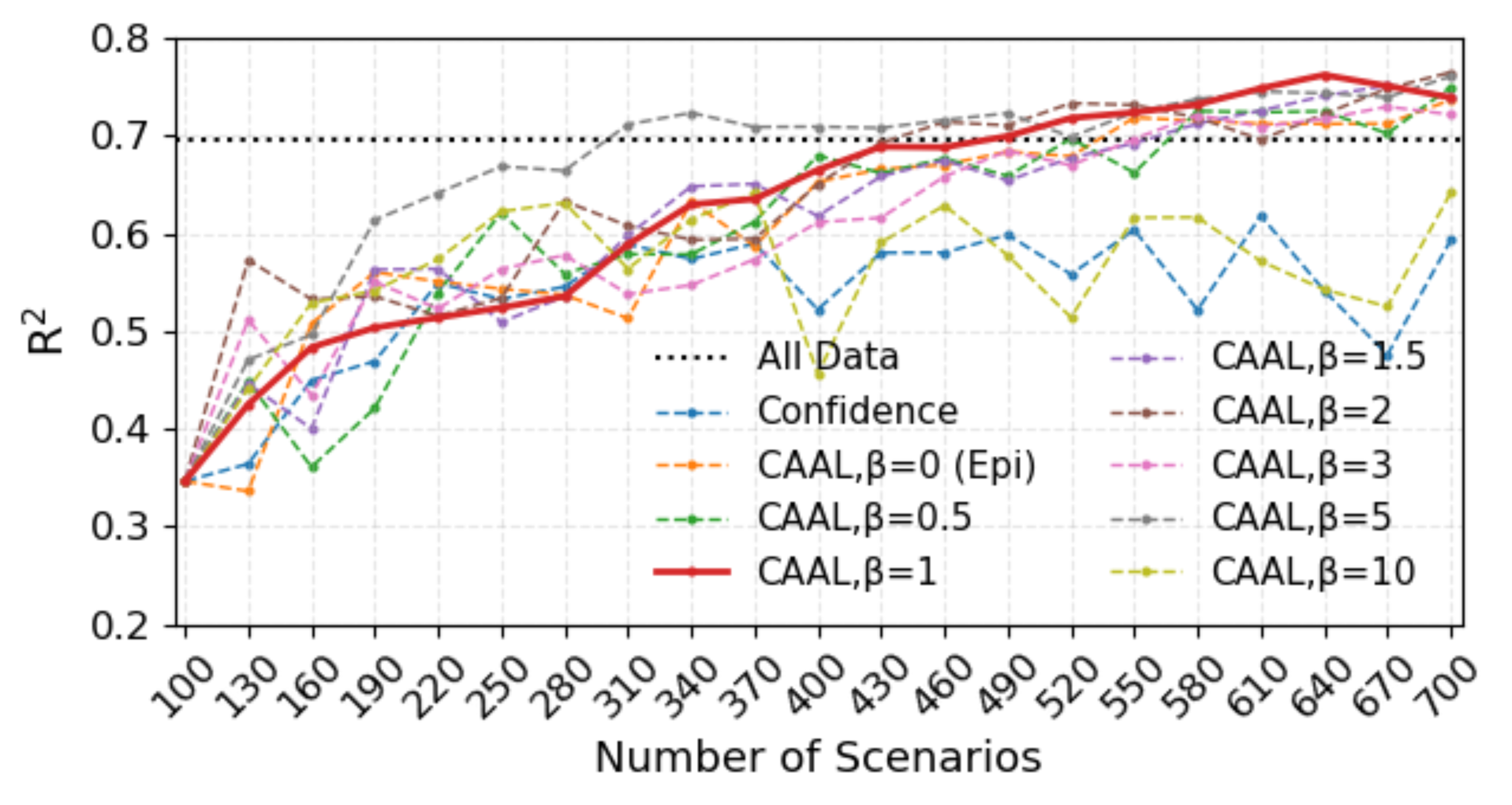}
  \caption{$R^2$ for $\chi_a$ on the PartMC dataset under CAAL with different $\beta$ values, evaluated at the scenario level.}
  \label{fig:beta_sensitivity}
\end{figure}

The hyperparameter $\beta$ controls the curvature of the reliability gate (confidence; $1-\tilde{\mathcal{U}}_{\mathrm{ale}}(\mathbf{x})$), effectively modulating the strategy's tolerance for noise. Figure~\ref{fig:beta_sensitivity} illustrates the performance trade-off.

\textbf{Gating disabled ($\beta = 0$).}
At this limit, the gating mechanism is disabled and the strategy reverts to pure epistemic sampling. Without suppressing aleatoric noise, it tends to select samples with high uncertainty regardless of their reliability, allowing noise-dominated points to enter the labelled set and thereby degrading the effectiveness of AL (Table~\ref{tab:main_results} and Figure~\ref{fig:beta_sensitivity}).

\textbf{Concave relaxation ($0<\beta <1$).} In this regime, the gating term ($(1-\tilde{\mathcal{U}}_{\mathrm{ale}}(\mathbf{x}))^{\beta}$) is concave, leading to a sub-linear penalty on aleatoric uncertainty. This acts as a mild noise filter: it suppresses extreme noise while remaining relatively permissive to moderate aleatoric levels, thereby maintaining exploration. In our experiments, however, this relaxation is not sufficiently selective. Some high-aleatoric samples still penetrate the filter, which degrades performance ($\beta=0.5$ in Figure~\ref{fig:beta_sensitivity}).

\textbf{Convex restriction ($\beta \geq 1$).} When $\beta>1$, the geometric profile of the gate flips to convex, imposing a super-linear penalty on noise to prioritise data cleanliness. In our case, this regime yields superior and stable CAAL performance for $\beta$ in the range of around $[1,5]$. However, as $\beta$ is set too large ($\beta \gg 1$), the gate becomes overly strict. Samples that are informative yet intrinsically noisy are excessively suppressed and therefore rarely queried. This reduces the diversity and difficulty of selected data, slowing learning and resulting in a noticeable drop in $R^2$ ($\beta=10$ in Figure~\ref{fig:beta_sensitivity}).

\subsection{Effect of Partial Loss Decoupling}
\label{sec:lambda_impact}

\begin{figure}
  \centering
  \includegraphics[width=\linewidth]{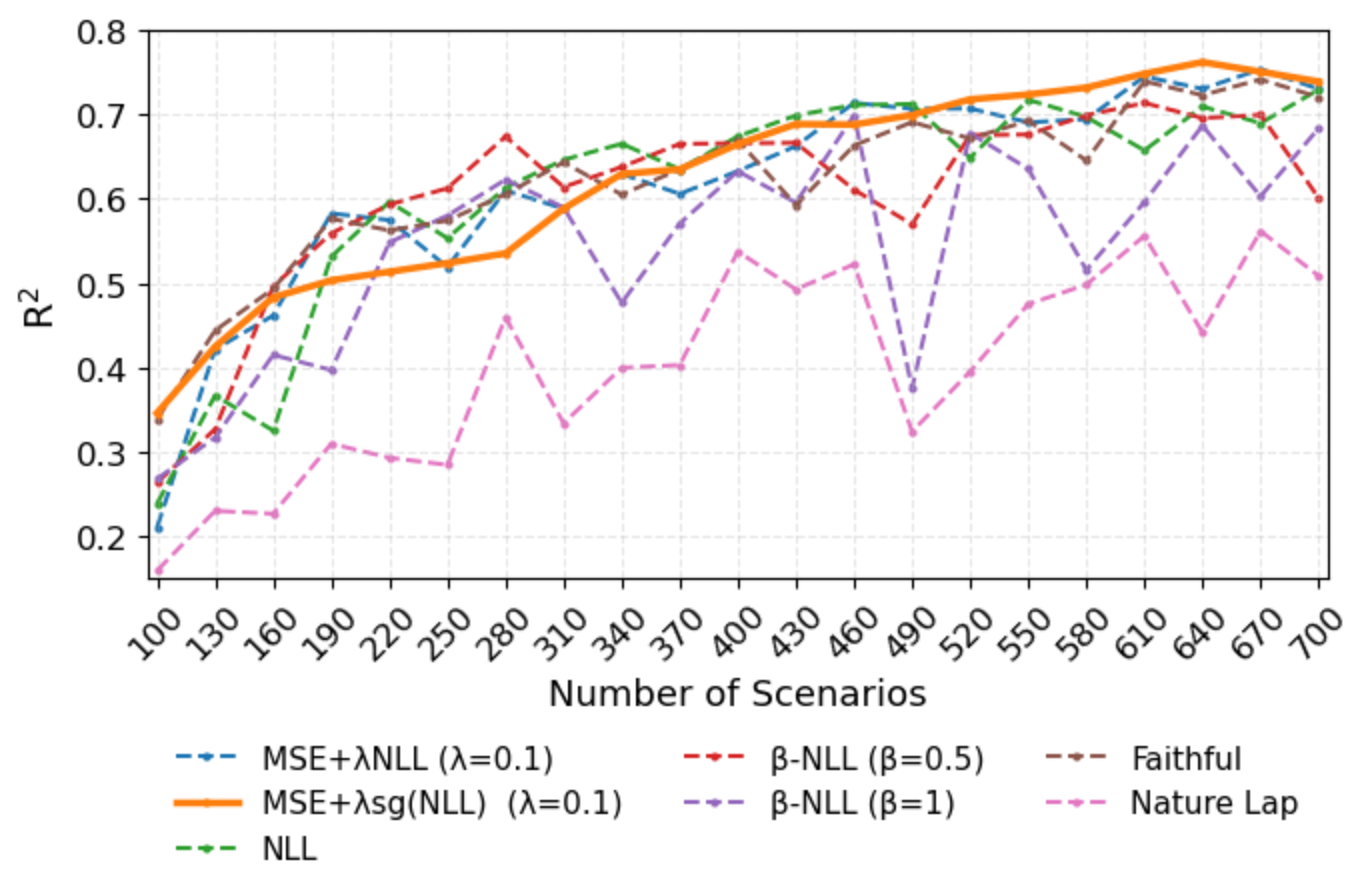}
  \caption{Comparison of training objectives under CAAL for predicting $\chi_a$ on the PartMC dataset.}
  \label{fig:loss_ab}
\end{figure}

\begin{figure}
  \centering
  \includegraphics[width=\linewidth]{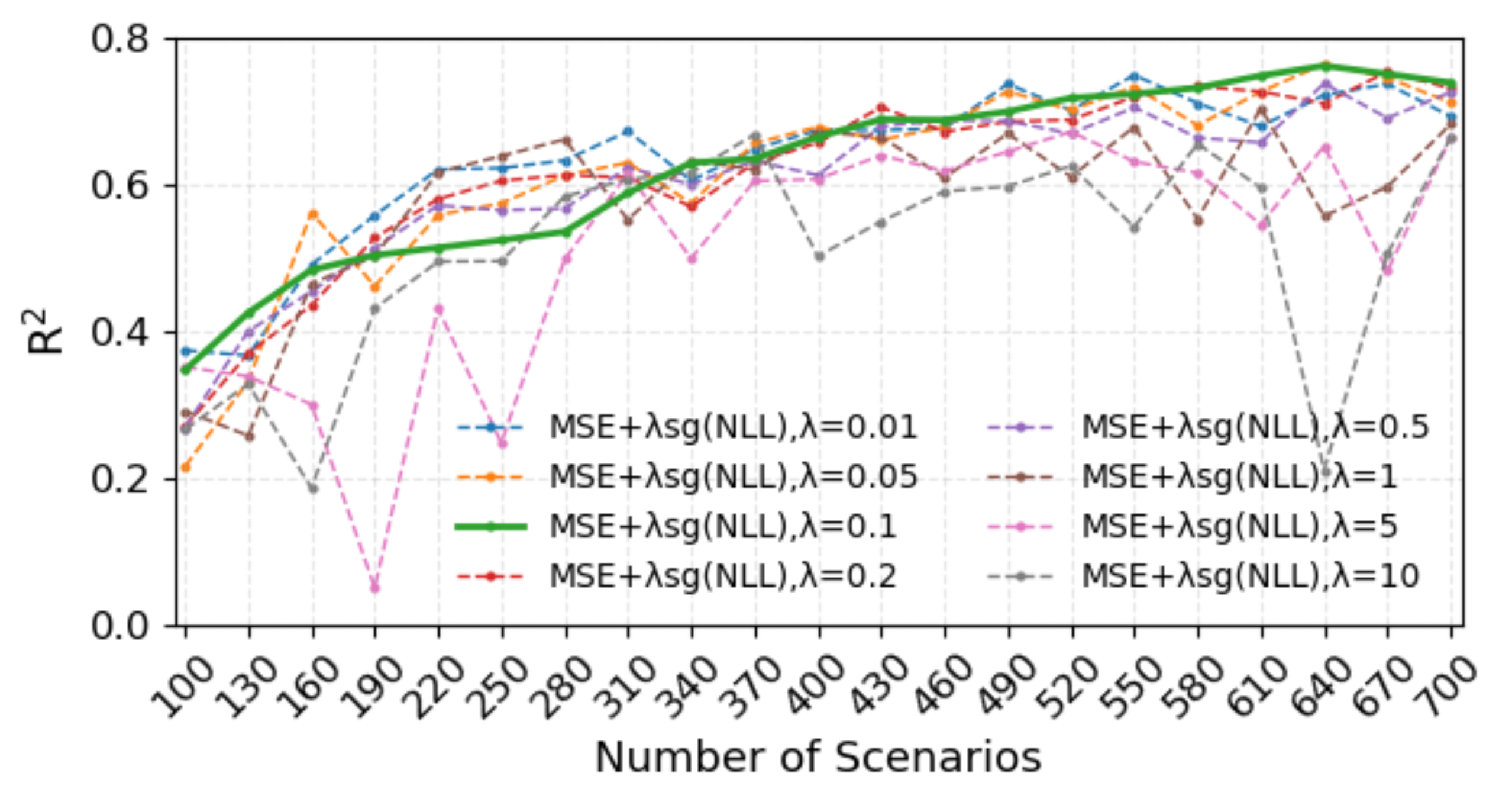}
  \caption{Sensitivity to the loss balancing coefficient $\lambda$ under CAAL for predicting $\chi_a$ on the PartMC dataset.}
  \label{fig:loss_lambda}
\end{figure}

\begin{table*}[t]
\centering
\caption{Best RMSE and $R^2$ achieved by different training objectives for $\chi_a$ on the PartMC dataset, reported separately for each AL strategy.}
\label{tab:loss_ablation}
\resizebox{\textwidth}{!}{
\begin{tabular}{l|cc|cc|cc|cc|cc|cc|cc|cc|cc|cc}
\hline
\textbf{Training Objectives} 
& \multicolumn{2}{c|}{\textbf{CAAL (Ours)}} 
& \multicolumn{2}{c|}{BALD} 
& \multicolumn{2}{c|}{LCMD} 
& \multicolumn{2}{c|}{BADGE} 
& \multicolumn{2}{c|}{Coreset} 
& \multicolumn{2}{c|}{ALM (Entropy)} 
& \multicolumn{2}{c|}{QBC (Epi)} 
& \multicolumn{2}{c|}{Random} 
& \multicolumn{2}{c|}{Ale} 
& \multicolumn{2}{c}{Confidence} \\
\cline{2-21}

& RMSE & $R^2$ 
& RMSE & $R^2$ 
& RMSE & $R^2$ 
& RMSE & $R^2$ 
& RMSE & $R^2$ 
& RMSE & $R^2$ 
& RMSE & $R^2$ 
& RMSE & $R^2$ 
& RMSE & $R^2$ 
& RMSE & $R^2$ \\
\hline

NLL \cite{NIXWEIGEND94Proc.1994IEEEInt.Conf.NeuralNetw.ICNN94}
& 0.0684 & 0.7295 
& 0.0680 & 0.7325 
& 0.0703 & 0.7148 
& 0.0711 & 0.7079 
& 0.0701 & 0.7157 
& 0.0710 & 0.7091 
& 0.0696 & 0.7207 
& 0.0684 & 0.7297 
& 0.0725 & 0.6968 
& 0.0809 & 0.6216 \\

MSE + $\lambda$NLL ($\lambda$=0.1) 
& 0.0654 & 0.7537 
& 0.0676 & 0.7363 
& 0.0700 & 0.7196 
& 0.0739 & 0.6849 
& 0.0690 & 0.7247 
& 0.0709 & 0.7101 
& 0.0688 & 0.7271 
& 0.0703 & 0.7148 
& 0.0709 & 0.7098 
& 0.0796 & 0.6339 \\

$\beta$-NLL ($\beta=0.5$) \cite{SEITZERETAL22}
& 0.0704 & 0.7137 
& \textbf{0.0670} & \textbf{0.7406} 
& \textbf{0.0690} & \textbf{0.7246} 
& \textbf{0.0691} & \textbf{0.7274}
& 0.0726 & 0.6955 
& 0.0718 & 0.7022 
& 0.0671 & 0.7404 
& 0.0687 & 0.7275 
& 0.0725 & 0.6969 
& 0.0782 & 0.6469 \\

$\beta$-NLL ($\beta=1$) \cite{SEITZERETAL22}
& 0.0722 & 0.6991 
& 0.0723 & 0.6982 
& 0.0736 & 0.6875 
& 0.0767 & 0.6605 
& 0.0693 & 0.7226 
& 0.0721 & 0.6995 
& 0.0706 & 0.7118 
& 0.0702 & 0.7155 
& \textbf{0.0709} & \textbf{0.7100} 
& 0.0836 & 0.5963 \\

Faithful \cite{STIRNETAL22}
& 0.0669 & 0.7416 
& 0.0684 & 0.7296 
& 0.0734 & 0.6887 
& 0.0694 & 0.7220 
& 0.0700 & 0.7170 
& 0.0711 & 0.7079
& \textbf{0.0670} & \textbf{0.7409} 
& 0.0703 & 0.7148 
& 0.0715 & 0.7052 
& \textbf{0.0773} & \textbf{0.6548} \\

Natural Lap \cite{Alexander23} 
& 0.0871 & 0.5618 
& 0.0826 & 0.6059 
& 0.0757 & 0.6689 
& 0.0722 & 0.6993 
& 0.0758 & 0.6684 
& 0.0857 & 0.5759 
& 0.0832 & 0.6005 
& 0.0772 & 0.6561 
& 0.0819 & 0.6124 
& 0.0820 & 0.6114 \\

\textbf{MSE + $\lambda$sg(NLL) ($\lambda=0.1$) (Ours)} 
& \textbf{0.0642} & \textbf{0.7621} 
& \textbf{0.0670} & \textbf{0.7406} 
& 0.0742 & 0.6817 
& 0.0700 & 0.7168 
& \textbf{0.0685} & \textbf{0.7291} 
& \textbf{0.0691} & \textbf{0.7248} 
& 0.0675 & 0.7365 
& \textbf{0.0682} & \textbf{0.7313} 
& 0.0711 & 0.7041 
& 0.0814 & 0.6180 \\

\hline
\end{tabular}
}
  \label{tab:loss_AL}
\end{table*}

Figure~\ref{fig:loss_ab} compares model performance under different training objectives. When optimising the NLL alone, CAAL exhibit lower $R^2$ and large performance fluctuations in the early stages of the AL process, and none of the acquisition strategies achieves the best overall performance under the NLL objective (Table~\ref{tab:loss_AL}). Introducing reweighting on top of NLL, such as the $\beta$-NLL loss with $\beta=0.5$, or adding an MSE term (MSE+$\lambda$NLL, $\lambda=0.1$), improves the stability of the AL process. However, these objectives still optimise the mean and variance jointly under an NLL-form term, where the residual is scaled by the predicted variance. As a result, both stability and final performance remain inferior to CAAL. At the other extreme, the Faithful approach completely blocks gradient propagation from the variance branch to both the mean head and the shared feature extractor. While this prevents noise-related gradients from corrupting mean and representation learning, it also suppresses the model’s ability to adapt to heteroscedastic structure in the data. In this task, overly strict isolation limits effective noise awareness and leads to weaker overall performance (Figure~\ref{fig:loss_ab}).

CAAL adopts a partially decoupled design that separates mean learning from aleatoric calibration, while retaining a controlled gradient pathway from the variance branch to the shared representation. This training form is well-suited to AL settings that use both epistemic and aleatoric uncertainty, beacuse the acquisition rule needs not only to identify informative regions but also to avoid samples dominated by irreducible noise. As shown in Table~\ref{tab:loss_AL}, this objective consistently delivers strong performance across uncertainty-driven acquisition strategies, with the clearest gains observed for approaches that combine epistemic and aleatoric signals such as CAAL and BALD.

\textbf{The regularisation parameter $\lambda$ further controls the balance between prediction learning and uncertainty modelling.} When $\lambda$ is large, optimisation becomes dominated by the uncertainty term, encouraging variance inflation rather than refinement of mean predictions and leading to degraded $R^2$ (Figure~\ref{fig:loss_lambda}). In contrast, CAAL maintains stable performance over a broad range of $\lambda$ values ($\lambda \in [0.01, 0.5]$), indicating low sensitivity to hyperparameter tuning. Within this range, the aleatoric loss term acts as an adaptive weighting mechanism that down-weights noisy samples without overwhelming the primary prediction objective.

\subsection{Robustness across Different Datasets}

\begin{table}
\centering
\caption{Comparison of effectiveness and efficiency across two tasks and data sources, reporting observational VR results first and simulation $\chi_o$ results second.}
\label{tab:main_results_stacked}
\resizebox{\linewidth}{!}{
\begin{tabular}{lcccccc}
\toprule
\textbf{Method} & Best $R^2$ ($\uparrow$) & vs. Full Data & Best RMSE ($\downarrow$) & vs. Full Data & Data to Match & \textbf{Labeling Saved} \\
\midrule

\textit{Full Data (VR)} & \textit{0.7102} & \textit{-} & \textit{2.4547} & \textit{-} & \textit{100\%} & \textit{0\%} \\
\midrule
Confidence & 0.7011 & $-1.28\%$ & 2.493 & $+1.56\%$ &  N/A & N/A \\
Ale & 0.7152 & $+0.70\%$ & 2.433 & $-0.865\%$ & 92.58\% & 7.42\% \\
ALM (Entropy) & 0.7294 & $+2.70\%$ & 2.372 & $-3.38\%$ & 83.51\% & 16.49\% \\
QBC (Epi) & 0.7189 & $+1.23\%$ & 2.417 & $-1.524\%$ & \textbf{62.34\%} & \textbf{37.66\%} \\
BALD & 0.7143 & $+0.57\%$ & 2.437 & $-0.709\%$ & 65.34\% & 34.66\% \\
LCMD & 0.7227 & $+1.77\%$ & 2.401 & $-2.20\%$ & 77.46\% & 22.54\% \\
BADGE & 0.7196 & $+1.32\%$ & 2.415 & $-1.64\%$ & 65.34\% & 34.66\% \\
Coreset & 0.7267 & $+2.33\%$ & 2.384 & $-2.89\%$ & \textbf{62.34\%} & \textbf{37.66\%} \\
Random & 0.7228 & $+1.78\%$ & 2.400 & $-2.21\%$ & 71.41\% & 28.59\% \\
\textbf{CAAL (Ours)} & \textbf{0.7343} & \textbf{+3.40\%} & \textbf{2.350} & \textbf{-4.26\%} & \textbf{62.34\%} & \textbf{37.66\%} \\
\midrule
\multicolumn{7}{l}{\textit{}} \\
\midrule
\textit{Full Data ($\chi_{o}$)} & \textit{0.7117} & \textit{-} & \textit{0.0775} & \textit{-} & \textit{100\%} & \textit{0\%} \\
\midrule
Confidence & 0.6789 & $-4.61\%$ & 0.0818 & $+5.63\%$ &  N/A & N/A \\
Ale & 0.7030 & $-1.22\%$ & 0.0787 & $+1.59\%$ &  N/A & N/A \\
ALM (Entropy) & 0.7063 & $-0.75\%$ & 0.0782 & $+1.019\%$ &  N/A & N/A \\
QBC (Epi) & 0.7128 & $+0.15\%$ & 0.0774 & $-0.09\%$ & 77.78\% & 22.22\% \\
BALD & 0.6939 & $-2.50\%$ & 0.0799 & $+3.13\%$ &  N/A & N/A \\
LCMD & 0.6728 & $-5.47\%$ & 0.0826 & $+6.64\%$ &  N/A & N/A \\
BADGE & 0.6706 & $-5.78\%$ & 0.0829 & $+7.00\%$ &  N/A & N/A \\
Coreset & 0.7094 & $-0.32\%$ & 0.0778 & $+0.49\%$ &  N/A & N/A \\
Random & 0.6923 & $-2.73\%$ & 0.0801 & $+3.41\%$ & N/A & N/A \\
\textbf{CAAL (Ours)} & \textbf{0.7198} & \textbf{+1.31\%} & \textbf{0.0764} & \textbf{-1.32\%} & \textbf{67.78\%} & \textbf{32.22\%} \\
\bottomrule
\end{tabular}}
\label{tab:stable}
\end{table}

Table~\ref{tab:stable} reports additional results under alternative evaluation settings, including predictions of the optical mixing state index $\chi_{optical}$ on the same PartMC simulations and predictions of VR using long-term Beijing observations.
These experiments are not intended as primary case studies, but are used to assess the robustness of CAAL to changes in target definitions and data sources.
Overall, the consistent gains across these settings suggest that CAAL does not depend on a specific mixing state formulation or a particular dataset.

\section{Conclusion}
We propose a Confidence-Aware Active Learning (CAAL) framework for heteroscedastic regression. On the training side, CAAL learns the predictive mean and variance separately using MSE and an NLL-based loss, with a single coefficient controlling the impact of variance learning on shared features. On the acquisition side, CAAL gates epistemic uncertainty using aleatoric uncertainty as a reliability signal to avoid noise-dominated queries. Experiments on PartMC simulations and observations show that CAAL achieves higher sample efficiency and more stable training than standard AL baselines. Overall, this work provides a practical pathway for constructing larger and higher-quality databases of high-cost atmospheric particle properties, laying a data foundation for stronger predictive models in environmental/health impact assessment.

\section{Acknowledgments}
This work is supported by The Aerosol Society Aerosol Science Career Development Grant.
\bibliographystyle{ACM-Reference-Format}
\bibliography{references}

\section{Appendices}

\appendix


\section{PartMC Dataset}
\label{append:partmc}
\subsection{Particle-Resolved Aerosol Model.}
PartMC-MOSAIC (Particle-resolved Monte Carlo code for atmospheric aerosol simulation) is a stochastic particle-resolved aerosol model \cite{RIEMERETAL09J.Geophys.Res.}. PartMC is a Lagrangian box model that tracks aerosol evolution in a well-mixed volume by stochastically simulating emission, coagulation, and dilution. Gas-phase chemistry and gas-aerosol partitioning are treated deterministically with MOSAIC, including CBM-Z for photochemistry \cite{zaveri_new_1999}, MTEM and MESA for inorganic aerosol thermodynamics\cite{ZAVERIETAL05J.Geophys.Res.}, and SORGAM for secondary organic aerosol formation \cite{SCHELLETAL01J.Geophys.Res.}. Model details are provided in \citeauthor{RIEMERETAL09J.Geophys.Res., DEVILLEETAL11AnnualRev.Anal.Chem., DEVILLEETAL19J.Comput.Dyn.} for PartMC, and \citeauthor{ZAVERIETAL08J.Geophys.Res.} for MOSAIC.

PartMC-MOSAIC is used to generate the dataset for the active learning (AL) project. Each PartMC-MOSAIC scenario simulates aerosol evolution under a distinct set of input parameters. These parameters, listed in Table~\ref{fig:table_2}, cover primary emissions of multiple aerosol types such as carbonaceous aerosol, sea salt, and dust across Aitken, accumulation, and coarse size ranges, primary emissions of gas phase species such as $SO_2$, $NO_2$, $CO$, and volatile organic compounds, and meteorological conditions. Parameter combinations for each scenario are selected using Latin Hypercube Sampling (LHS).

\subsection{Aerosol Mixing State Definition.}
Atmospheric aerosols are suspensions of solid or liquid particles in air, spanning a wide range of sizes and chemical compositions \cite{RAESETAL00AtmosphericEnvironment, POSCHL05AngewChemIntEd}. Aerosol mixing state describes how chemical composition varies across particles within an aerosol population \cite{WINKLER73JournalofAerosolScience, RIEMERETAL19ReviewsofGeophysics}. In the fully internally mixed case, all particles share the same composition. In the fully externally mixed case, each particle consists of only one chemical species \cite{RIEMERETAL19ReviewsofGeophysics}. Most ambient aerosols fall between these two idealised extremes \cite{HEALYETAL14AtmosphericChem.Phys., BONDYETAL18Atmos.Chem.Phys., LEEETAL19Environ.Sci.Technol.}. To quantify this variability, \citeauthor{RIEMERWEST13Atmos.Chem.Phys.} defined the aerosol mixing state index, $\chi$, based on Shannon entropy of the distribution of chemical species among particles. $\chi$ ranges from 0$\%$ (completely externally mixed) to 100$\%$ (completely internally mixed), with larger values indicating more similar particle compositions and a higher degree of internal mixing.

\subsection{Aerosol Mixing State Index Calculations.}
The mixing state index $\chi$ is given by the affine ratio of the average particle species diversity, $D_{\alpha}$, and bulk population species diversity, $D_{\gamma}$, as
\begin{equation} \label{eq:mixing_state_index}
 {\chi} = \frac{D_{\alpha}-1}{D_{\gamma}-1}. 
\end{equation}

The following are the calculations for the diversities $D_{\alpha}$ and $D_{\gamma}$. First, the per-particle mixing entropies $H_i$ are calculated for each particle by

\begin{equation} \label{eq:entropy}
H_i = - \sum^{A}_{a=1} p_{i}^{a} \ln{p_{i}^{a}},
\end{equation}
\noindent
where $A$ is the number of distinct aerosol species and $p_i^a$ is the mass fraction of species $a$ in particle $i$. These values are then averaged (mass-weighted) over the entire population to obtain the average particle species diversity $D_{\alpha}$ by

\begin{equation} \label{eq:H_alpha}
H_{\alpha} = \sum^{N_{\rm p}}_{i=1} p_{i} H_{i}, 
\end{equation}

\begin{equation} \label{eq:D_alpha}
D_\alpha = e^{H_\alpha},
\end{equation}
\noindent
where $N_{\rm p}$ is the total number of particles in the population and $p_i$ is the mass fraction of particle $i$ in the population. Finally, the bulk diversity $D_\gamma$ is calculated as

\begin{equation} 
H_{\gamma} = -\sum^{A}_{a=1} p^{a} \ln{p^a}, \label{eq:H_gamma}
\end{equation}

\begin{equation} 
D_{\gamma}=e^{H_{\gamma}}, \label{eq:D_gamma}
\end{equation}
\noindent
where $p^{a}$ is the bulk mass fraction of species $a$ in the population. 

We compute two types of $\chi$ from the same simulation dataset, the chemical species mixing state index $\chi_a$ and the optical mixing state index $\chi_o$ \cite{ZHENGETAL21EarthSpaceSci.}. The $\chi_a$ index is derived from the abundance of the major aerosol chemical species. For $\chi_o$, black carbon is treated as the optically absorbing component, while all remaining aerosol species are grouped and assumed to be non-absorbing. The definitions of both indices are summarised in Table~\ref{tab:chi_base}, and the model input features are listed in Table~\ref{tab:chi_input}.

Since the mixing state index $\chi$ is bounded in $[0,1]$, we transform the target into an unbounded space using a logit mapping before training:
\begin{equation}
z = \log\frac{\chi}{1-\chi}.
\end{equation}
To avoid numerical instability, $\chi$ is first clipped to $[\varepsilon, 1-\varepsilon]$ with $\varepsilon = 10^{-6}$. 
The heteroscedastic NLL loss is then optimised in the logit space, and the predicted mean is mapped back to $[0,1]$ using the sigmoid function at inference time.

\begin{table}[H]
    \centering
\caption{Key aerosol species (size range: 100--700 nm) used in calculating the mixing state indexes $\chi_a$ and $\chi_o$ for PartMC simulations.}
\label{tab:MixingState}
    \begin{tabular}{cc}
    \hline
         Aerosol mixing state index & Group species\\
         \hline
         Abundance ($\chi_a$) & [BC], [SO$_4$], [NO$_3$], [NH$_4$], [OA]\\
         Optical property ($\chi_o$) & [BC], [SO$_4$, NO$_3$, NH$_4$, OA]\\
         \hline
    \end{tabular}
    \label{tab:chi_base}
\end{table}

\begin{table}[H]
    \centering
\caption{Model input feature subsets of $\chi_a$ and $\chi_o$ for PartMC simulations.}
\label{tab:MixingState}
    \begin{tabular}{ccc}
    \hline
         Input feature subsets & Features & Unit\\
         \hline
          & organic aerosols (OA)& $\mu g/m^3$\\
          & black carbon (BC)& $\mu g/m^3$\\
          Aerosol& nitrate (NO$_3$)& $\mu g/m^3$\\
          & sulfate (SO$_4$)& $\mu g/m^3$\\
          & ammonium (NH$_4$)& $\mu g/m^3$\\
          \hline
          Environment & temperature (T)& K\\
          & relative humidity (RH)& \\
          \hline
          & carbon monoxide (CO)& ppb\\
          Non--VOC gas & nitrogen oxides (NO$_x$)& ppb \\
          & nitrous oxides (NO)& ppb \\
          & ozone (O$_3$)& ppb \\
        \hline
          & xylene (XYL)& ppb\\
          & ethene (ETH) & ppb \\
          & acetone (AONE) & ppb \\
          VOC gas& toluene (TOL)& ppb \\
          & paraffin carbon (PAR) & ppb \\
          & internal olefin carbons (OLET) & ppb \\
          & acetaldehyde (ALD2) & ppb \\
         \hline
    \end{tabular}
    \label{tab:chi_input}
\end{table}

\begin{table*}
\caption{List of input parameters and their sampling ranges to construct the training and testing scenarios (particle-resolved simulation). The variables $D_{\rm
  g}$, $\sigma_{\rm g}$, $E_{\rm a}$ refer to geometric mean
 diameter, geometric standard deviation, and number emission flux,
 respectively. }\label{tab:para_defs}
\begin{tabular}{ll}
\hline
\textbf{Parameters}  & \textbf{Range} \\
\hline
\textbf{Environmental Variable} & \\
Relative humidity (RH) & [0.1, 1) or [0.4, 1) \\
Latitude & (70$^\circ$S, 70$^\circ$N) or (90$^\circ$S, 90$^\circ$N)\\
Day of Year & [1, 365] \\
Temperature & Varies with time of day and location\textsuperscript{1} \\
\hline
\textbf{Gas Phase Emissions Scaling Factor} & \\
SO$_{2}$, NO$_{2}$, NO, NH$_{3}$, CO, CH$_{3}$OH, &\\
ALD2 (Acetaldehyde), ANOL (Ethanol), &\\
AONE (Acetone), DMS (Dimethyl sulfide), &\\
ETH (Ethene), HCHO (Formaldehyde), & [0, 200${\%}$] of the reference scenario\\
ISOP (Isoprene), OLEI (Internal olefin carbons), &\\
OLET (Terminal olefin carbons), &\\
PAR (Paraffin carbon), TOL (Toluene), XYL (Xylene) &\\
\hline
\textbf{Carbonaceous Aerosol Emissions (one mode)} & \\
$D_{\rm {g}}$ & [25 nm, 250 nm] \\
$\sigma_{\rm {g}}$ & [1.4, 2.5] \\
BC/OC mass ratio & [0, 100${\%}$] \\
$E_{\rm {a}}$ & [0, 1.6 $\times$ 10$^7$ m$^{-2}$ s$^{-1}$] \\
\hline
\textbf{Sea Salt Emissions (two modes)}& \\
$D_{\rm {g,1}}$ & [180 nm, 720 nm] \\
$\sigma_{\rm {g,1}}$ & [1.4, 2.5] \\
$E_{\rm {a,1}}$ & [0, 1.69 $\times$ 10$^5$ m$^{-2}$ s$^{-1}$] \\
$D_{\rm {g,2}}$ & [1 {$\mu$}m, 6 {$\mu$}m] \\
$\sigma_{\rm {g,2}}$ & [1.4, 2.5] \\
$E_{\rm {a,2}}$ & [0, 2380 m$^{-2}$ s$^{-1}$] \\
OC fraction & [0, 20${\%}$]\\
\hline
\textbf{Dust Emissions (two modes)}& \\
$D_{\rm {g,1}}$ & [80 nm, 320 nm] \\
$\sigma_{\rm {g,1}}$ & [1.4, 2.5] \\
$E_{\rm {a,1}}$ & [0, 5.86 $\times$ 10$^5$ m$^{-2}$ s$^{-1}$] \\
$D_{\rm {g,2}}$ & [1 {$\mu$}m, 6 {$\mu$}m] \\
$\sigma_{\rm {g,2}}$ & [1.4, 2.5] \\
$E_{\rm {a,2}}$ & [0, 2380 m$^{-2}$ s$^{-1}$] \\
\hline
\textbf{Restart Timestamp}& \\
Timestamp & [0, 24 hours]\\
\hline
\begin{footnotesize}
\end{footnotesize}

\end{tabular}

\label{fig:table_2}
\end{table*}

\section{Observational Dataset}
\label{append:VR}
\subsection{Black Carbon Coating Definition.} Black carbon (BC) is a major light-absorbing aerosol produced primarily by incomplete combustion and can influence climate and air quality \cite{BONDETAL13JGRAtmospheres, HUANGETAL23Science}. 
Atmospheric BC-containing particles (BCc) typically comprise a refractory BC core (rBC) mixed with non-rBC materials that form coatings. 
The relative amount of coating is quantified by the coating volume ratio (VR), defined as the ratio of the coating volume to the rBC volume in BCc.
\begin{equation}
\mathrm{VR}=\frac{V_{\mathrm{coating}}}{V_{\mathrm{rBC}}}.
\end{equation}
Larger VR indicates a greater fraction of non-rBC material relative to rBC, which is closely linked to BC optical properties \cite{LIUETAL17NatureGeosci, JACOBSON01Nature, PENGETAL16Proc.Natl.Acad.Sci.U.S.A.}.

\subsection{Observation-Based Derivation of BC Coating Parameters.}
We derive the VR of BCc from observations using a single-particle soot photometer (SP2; DMT Inc.) \cite{SCHWARZETAL06J.Geophys.Res.}. The SP2 uses an intracavity Nd:YAG laser (1064\~nm) to heat rBC to incandescence \cite{SLOWIKETAL07AerosolScienceandTechnology}. The incandescence signal is approximately proportional to rBC mass and is used to retrieve the rBC core equivalent volume diameter, denoted as $D_c$. The SP2 also measures the elastic scattering signal of individual BCc particles. To mitigate laser-induced volatilisation effects, we fit only the leading edge of the scattering signal. The corrected scattering amplitude is then matched to a Mie-theory lookup table to infer the coated particle diameter $D_p$ \cite{LIUETAL14Atmos.Chem.Phys., MOTEKIETAL10JournalofAerosolScience, TAYLORETAL15Atmos.Meas.Tech.}.

For a time interval $\Delta t$, suppose $N$ BCc particles are detected, indexed by $i=1,\ldots,N$. Assuming spherical-equivalent volumes, we define the population-mean coating volume ratio as
\begin{equation}
\mathrm{VR}(\Delta t)
= \frac{\sum_{i=1}^{N} D_{p,i}^{3}}{\sum_{i=1}^{N} D_{c,i}^{3}} - 1,
\end{equation}
where $D_{p,i}$ and $D_{c,i}$ denote the coated particle diameter and the rBC core diameter of the $i$-th particle, respectively.

Since we only consider BC-containing samples, the VR is strictly positive. We therefore model the target in log-space and apply a Gaussian likelihood to $\log(\mathrm{VR})$.

\begin{table}[H]
    \centering
\caption{Model input feature subsets of VR for Beijing Observations.}
\label{tab:MixingState}
    \begin{tabular}{ccc}
    \hline
         Input feature subsets & Features & Unit\\
         \hline
          Aerosol& black carbon (BC)& $\mu g/m^3$\\
          & Particulate Matter 2.5 (PM$_{2.5}$)& $\mu g/m^3$\\
          \hline
          Environment & temperature (T)& K\\
          & relative humidity (RH)& \\
          \hline
          & carbon monoxide (CO)& ppb\\
          Gas & nitrogen oxides (NO$_x$)& ppb \\
          & sulphur dioxide (SO$_2$)& ppb \\
          & ozone (O$_3$)& ppb \\
        \hline
    \end{tabular}
        \label{tab:vr_input}
\end{table}

\section{Model Configuration.}
\label{sec:config}
Models for predicting $\chi_a$ are trained using the PartMC simulation dataset, and hyperparameters are tuned with Optuna. The corresponding search space and ranges are summarised in Table~\ref{tab:hyper}. The same hyperparameter configuration is used for all models throughout this study. The only difference between configurations is the number of input features, which varies according to the selected feature sets.

The prediction of $\chi_a$ and $\chi_o$ uses the same set of input features, as listed in Table~\ref{tab:chi_input}. In contrast, the model for VR prediction adopts a different feature set, which is summarised in Table~\ref{tab:vr_input}.

\begin{table}[H]
    \centering
\caption{Hyperparameters used to train FT-transformer are listed along with their optimal values and
the search space}
\label{tab:MixingState}
    \begin{tabular}{ccc}
    \hline
         Hyperparameter & Value & Search\\
         \hline
          layers& 4& [2--6] (step = 1)\\
          feed-forward dimension & 512& [128,256,512,1024]\\
          batch size & 128 & [32,64,128]\\
          initial learning rate & 1e-4 & [1e-5 -- 1e-3] (log-uniform) \\
          attention heads & 8 & [2,4,8]\\
          hidden size & 128 & [64,128,256] \\
          weight decay & 1e-4 & [1e-5 -- 1e-3]  (log--uniform) \\
        \hline
    \end{tabular}
    \label{tab:hyper}
\end{table}

\section{Performance under the Full Simulation Data Regime}
\label{append:alldata}
In addition to the limited-budget setting used in the main experiments, we further evaluate all AL strategies on the PartMC simulation dataset under the full-data regime, where the labelled set is gradually expanded until the entire unlabeled pool is exhausted. The experiment consists of 26 acquisition rounds, and 30 scenarios are selected in each round.

\begin{figure*}
  \includegraphics[width=\linewidth]{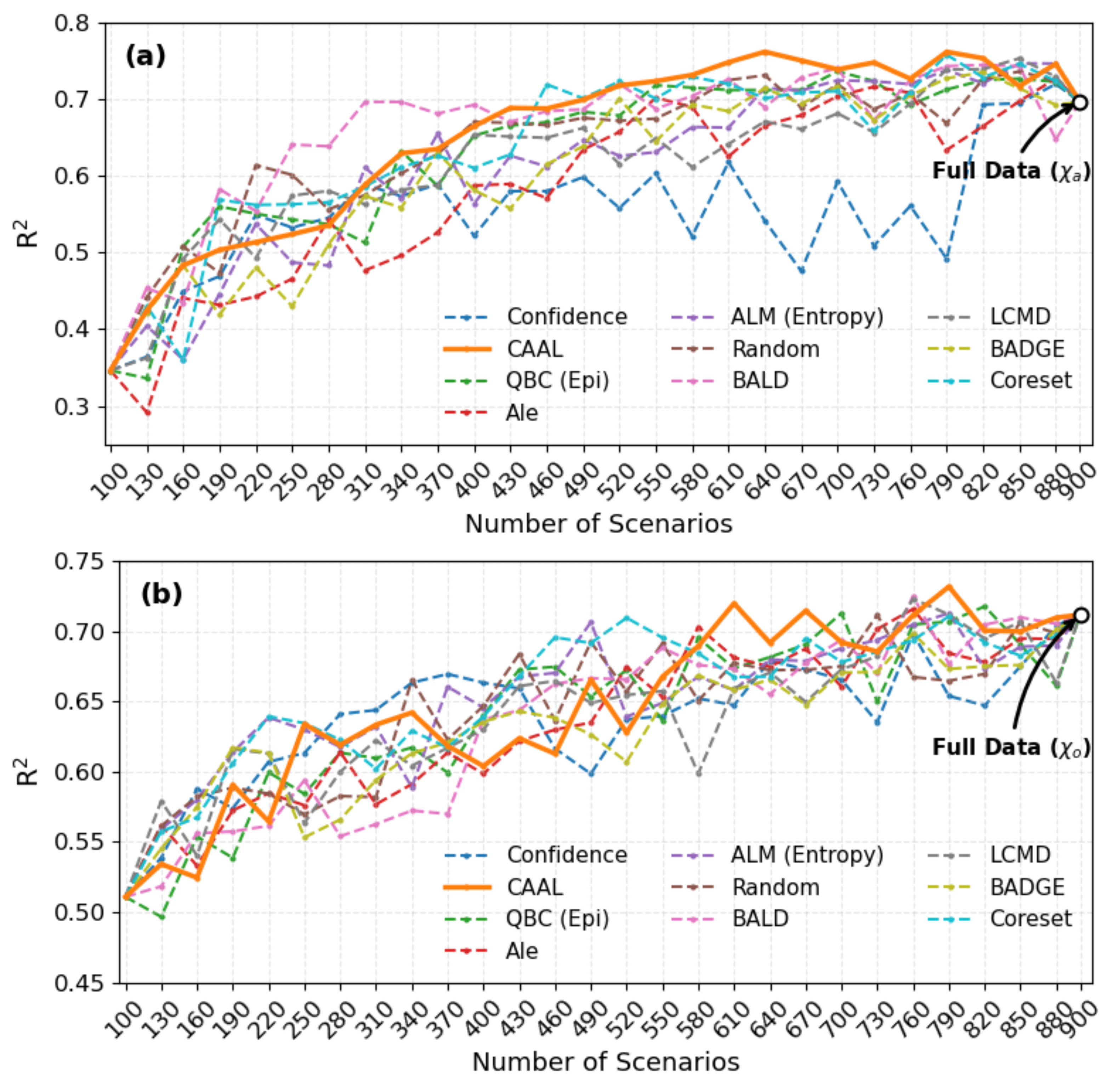}
  \caption{Learning curves of $R^2$ for (a) $\chi_a$ and (b) $\chi_o$ on the PartMC dataset as a function of the number of selected scenarios. Each curve corresponds to one sampling strategy. The black marker indicates the performance obtained when using the full training set.}
  \label{fig:r2_alldata}
\end{figure*}

Figure~\ref{fig:r2_alldata} and Table~\ref{tab:stable1} present the learning curves on the PartMC dataset for the prediction of $\chi_a$ and $\chi_o$. CAAL remains competitive across the entire acquisition process and does not degrade as the labelling budget approaches the full-data regime. This behaviour is consistent with the conclusions obtained under the limited budget setting in the main paper.

\begin{table*}
\centering
\caption{Performance comparison of different active learning strategies under the full-data regime on the PartMC simulation dataset. 
Results are reported for two prediction targets, $\chi_a$ and $\chi_o$. 
For each method, we report the best test $R^2$ and RMSE achieved during the acquisition process (26 AL rounds in total), together with the relative change with respect to training on the full PartMC labelled dataset. 
``Data to Match'' denotes the fraction of labelled scenarios required to reach the full-data performance, and ``Labelling Saved'' indicates the corresponding reduction in labelling cost.}
\label{tab:main_results_stacked}
\resizebox{\linewidth}{!}{
\begin{tabular}{lcccccc}
\toprule
\textbf{Method} & Best $R^2$ ($\uparrow$) & vs. Full Data & Best RMSE ($\downarrow$) & vs. Full Data & Data to Match & \textbf{Labeling Saved} \\
\midrule

\textit{Full Data ($\chi_{a}$)} & \textit{0.6966} & \textit{-} & \textit{0.0728} & \textit{-} & \textit{100\%} & \textit{0\%} \\
\midrule
Confidence & 0.7199 & $+3.34\%$ & 0.0697 & $-4.33\%$ &  97.8\% & 2.2\% \\
Ale & 0.7271 & $+4.38\%$ & 0.0687 & $-5.57\%$ & 61.1\% & 38.9\% \\
ALM (Entropy) & 0.7468 & $+7.21\%$ & 0.0662 & $-9.05\%$ & 71.1\% & 28.9\% \\
QBC (Epi) & 0.7365 & $+5.73\%$ & 0.0676 & $-7.21\%$ & 61.1\% & 38.9\% \\
BALD & 0.7450 & $+6.94\%$ & 0.0665 & $-8.71\%$ & 57.8\% & 42.2\% \\
LCMD & 0.7534 & $+8.16\%$ & 0.0653 & $-10.24\%$ & 87.8\% & 22.2\% \\
BADGE & 0.7354 & $+5.56\%$ & 0.0677 & $-7.01\%$ & 57.8\% & 42.2\% \\
Coreset & 0.7580 & $+8.82\%$ & 0.0646 & $-11.25\%$ & \textbf{51.1\%} & \textbf{48.9\%} \\
Random & 0.7363 & $+5.70\%$ & 0.0676 & $-7.18\%$ & 67.8\% & 32.2\% \\
\textbf{CAAL (Ours)} & \textbf{0.7621} & \textbf{+9.40\%} & \textbf{0.0642} & \textbf{-11.83\%} & 54.4\% & 45.6\% \\
\midrule
\multicolumn{7}{l}{\textit{}} \\
\midrule
\textit{Full Data ($\chi_{o}$)} & \textit{0.7117} & \textit{-} & \textit{0.0775} & \textit{-} & \textit{100\%} & \textit{0\%} \\
\midrule
Confidence & 0.7038 & $-1.10\%$ & 0.0786& $+1.45\%$ &  100\% &  0\% \\
Ale & 0.7157 & $+0.56\%$ & 0.0770 & $-0.60\%$ &  84.4\% & 15.6\% \\
ALM (Entropy) & 0.7127 & $+0.14\%$ & 0.0774 & $-0.08\%$ &  87.8\% & 12.2\% \\
QBC (Epi) & 0.7176 & $+0.82\%$ & 0.0767 & $-0.93\%$ & 77.8\% & 22.2\% \\
BALD & 0.7249 & $+1.86\%$ & 0.0757 & $-2.23\%$ &  84.4\% & 15.6\% \\
LCMD & 0.7230 & $+1.58\%$ & 0.0760 & $-1.88\%$ &  84.4\% & 15.6\% \\
BADGE & 0.7012 & $-1.47\%$ & 0.0789 & $+1.89\%$ &  100\% &  0\% \\
Coreset & 0.7105 & $-0.17\%$ & 0.0777 & $+0.30\%$ &  100\% &  0\% \\
Random & 0.7114 & $-0.05\%$ & 0.0776 & $+0.15\%$ & 100\% &  0\% \\
\textbf{CAAL (Ours)} & \textbf{0.7316} & \textbf{+2.80\%} & \textbf{0.0748} & \textbf{-3.43\%} & \textbf{67.78\%} & \textbf{32.22\%} \\
\bottomrule
\end{tabular}}
\label{tab:stable1}
\end{table*}

\begin{figure*}
  \includegraphics[width=\linewidth]{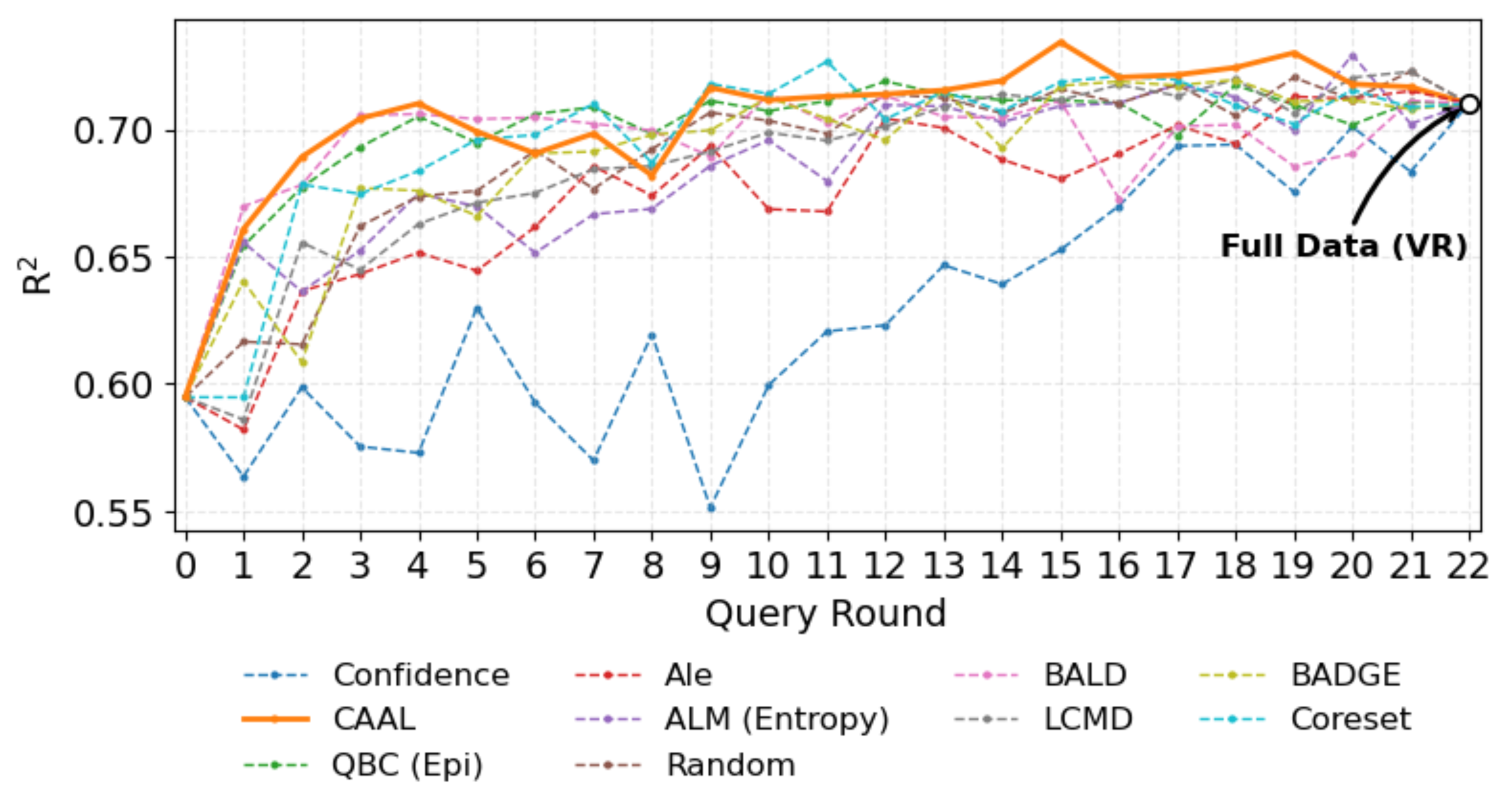}
  \caption{Learning curves of $R^2$ for VR on the Beijing observations as a function of the number of selected scenarios. Each curve corresponds to one sampling strategy. The black marker indicates the performance obtained when using the full training set.}
  \label{fig:vr_r2}
\end{figure*}

\begin{figure*}\includegraphics[width=\textwidth]{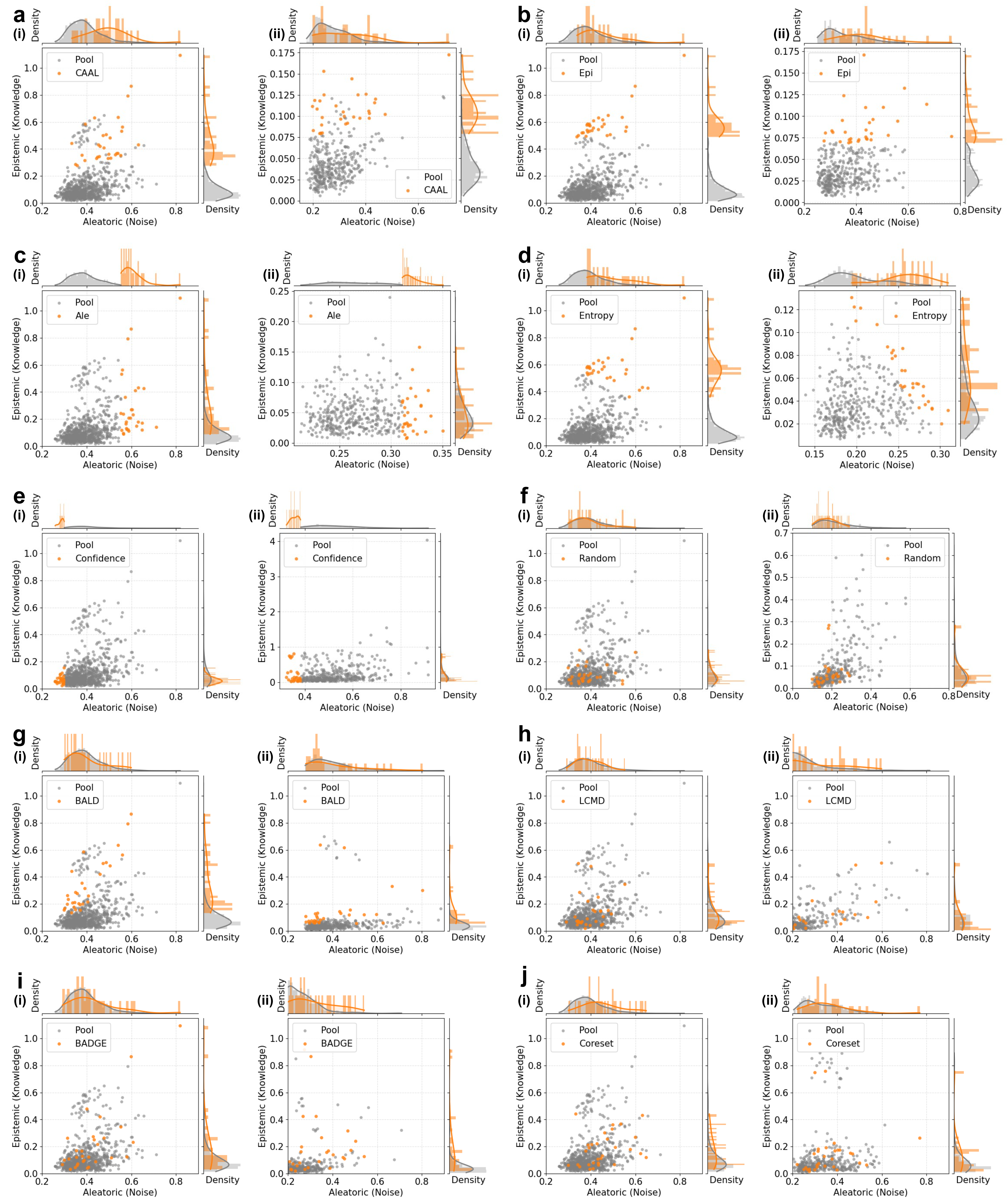}
  \caption{Joint distribution analysis of epistemic and aleatoric uncertainty for AL selection (a--j) in the prediction of $\chi_a$ on the PartMC dataset at round 1 (i) and round 14 (ii). The central scatter plots visualise the specific decision boundary in the uncertainty space, where orange points represent selected samples and grey points represent the unselected pool. The marginal histograms (top and right) quantify the distributional shift between selected and unselected sets.}
  \label{fig:epivs}
\end{figure*}

\end{document}